%% file: main.tex
\documentclass[utf8]{frontiersSCNS} 

\usepackage[USenglish]{babel}
\usepackage{graphicx} 
\graphicspath{{./images/}{./plots/}} 
\usepackage[]{mathtools} 
\usepackage[labelformat=parens]{subcaption} 
\usepackage[hang]{footmisc}
\usepackage[export]{adjustbox} 
\usepackage{wrapfig}
\usepackage{tabularx}
\usepackage{colortbl}
\usepackage{booktabs}
\usepackage{multirow}
\usepackage{csquotes}
\usepackage{pgf}

\usepackage{microtype}
\usepackage[onehalfspacing]{setspace}

\def\keyFont{\fontsize{8}{11}\helveticabold }
\def\firstAuthorLast{Muratore~{et~al.}}
\def\Authors{%
	\mbox{Fabio~Muratore$^{1,2,*}$,
	Fabio~Ramos$^{3,4}$,
	Greg~Turk$^{5}$,
	Wenhao~Yu$^{6}$,
	Michael~Gienger$^{2}$} and
	Jan~Peters$^{1}$
}
\def\Address{%
	$^{1}$\mbox{Intelligent Autonomous Systems Group, Technical University of Darmstadt, Germany}\\%
	$^{2}$Honda Research Institute Europe, Offenbach am Main, Germany\\%
	$^{3}$School of Computer Science, University of Sydney, Australia\\%
	$^{4}$NVIDIA, Seattle, WA, USA\\%
	$^{5}$Georgia Institute of Technology, Atlanta, GA, USA\\%
	$^{6}$Robotics at Google, Mountain View, CA, USA%
}
\def\corrAuthor{Fabio Muratore, fabio@robot-learning.de\vspace{-0.8\baselineskip}} 
\def\corrEmail{}

\usepackage{titlesec}
\titlespacing{\subsection}{0pt}{1.5ex plus 1ex minus 1ex}{*0}
\titlespacing{\paragraph}{0pt}{0.9ex plus 0.6ex minus 0.6ex}{*0}

\usepackage[lined,linesnumbered,ruled]{algorithm2e}

\usepackage[nolist,nohyperlinks]{acronym}
\IfFileExists{\string~/HESSENBOX-DA/fMRT_LaTeX/fMRT_acronyms}%
{\input{\string~/HESSENBOX-DA/fMRT_LaTeX/fMRT_acronyms}}
{\input{fMRT_acronyms}}
\IfFileExists{\string~/HESSENBOX-DA/fMRT_LaTeX/fMRT_macros}%
{\input{\string~/HESSENBOX-DA/fMRT_LaTeX/fMRT_macros}}
{\input{fMRT_macros}}

\usepackage[hidelinks,bookmarks=true,breaklinks]{hyperref} 

\begin{document}
\onecolumn
\firstpage{1}

\title[Robot Learning from Randomized Simulations]{Robot Learning from Randomized Simulations: A Review}

\author[\firstAuthorLast ]{\Authors} 
\address{} 
\correspondance{} 
\extraAuth{}

\maketitle

\begin{abstract}
\noindent The rise of deep learning has caused a paradigm shift in robotics research, favoring methods that require large amounts of data.
Unfortunately, it is prohibitively expensive to generate such data sets on a physical platform.
Therefore, \sota approaches learn in simulation where data generation is fast as well as inexpensive and subsequently transfer the knowledge to the real robot (\simtoreal).
Despite becoming increasingly realistic, all simulators are by construction based on models, hence inevitably imperfect.
This raises the question of how simulators can be modified to facilitate learning robot control policies and overcome the mismatch between simulation and reality, often called the \enquote*{reality gap}.
We provide a comprehensive review of \simtoreal research for robotics, focusing on a technique named \enquote*{domain randomization} which is a method for learning from randomized simulations.

\tiny
\keyFont{\section{Keywords:} robotics, simulation, reality gap, simulation optimization bias, reinforcement learning, domain randomization, sim-to-real} 
\end{abstract}

\section{Introduction}
Given that machine learning has achieved super-human performance in image classification~\citep{Ciresan_Schmidhuber_12,Krizhevsky_Hinton_12} and games \citep{Mnih_Hassabis_15,Silver_Hassabis_16}, the question arises why we do not see similar results in robotics.
There are several reasons for this.
First, learning to act in the physical world is orders of magnitude more difficult.
While the data required by modern (deep) learning algorithms could be acquired directly on a real robot~\citep{Levine_Quillen_16}, this solution is too expensive in terms of time and resources to scale up.
Alternatively, the data can be generated in simulation faster, cheaper, safer, and with unmatched diversity.
In doing so, we have to cope with unavoidable approximation errors that we make when modeling reality.
These errors, often referred to as the \enquote*{reality gap}, originate from omitting physical phenomena, inaccurate parameter estimation, or the discretized numerical integration in typical solvers.
Compounding this issue, \sota (deep) learning methods are known to be brittle \citep{Szegedy_Fergus_13,Goodfellow_Szegedy_14,Huang_Abbeel_17}, that is, sensitive to shifts in their input domains.
Additionally, the learner is free to exploit the simulator, overfitting to features which do not occur in the real world.
For example, \citet{Baker_Mordatch_20} noticed that the agents learned to exploit the physics engine to gain an unexpected advantage.
While this exploitation is an interesting observation for studies made entirely in simulation, it is highly undesirable in \simtoreal scenarios.
In the best case, the reality gap manifests itself as a performance drop, giving a lower success rate or reduced tracking accuracy.
More likely, the learned policy is not transferable to the robot because of unknown physical effects.
One effect that is difficult to model is friction, often leading to an underestimation thereof in simulation, which can result in motor commands that are not strong enough to get the robot moving.
Another reason for failure are parameter estimation errors, which can quickly lead to unstable system dynamics.
This case is particularly dangerous for the human and the robot.
For these reasons, bridging the reality gap is the essential step to endow robots with the ability to learn from simulated experience.

\begin{figure}[t]
	\centering
	\includegraphics[width=\linewidth]{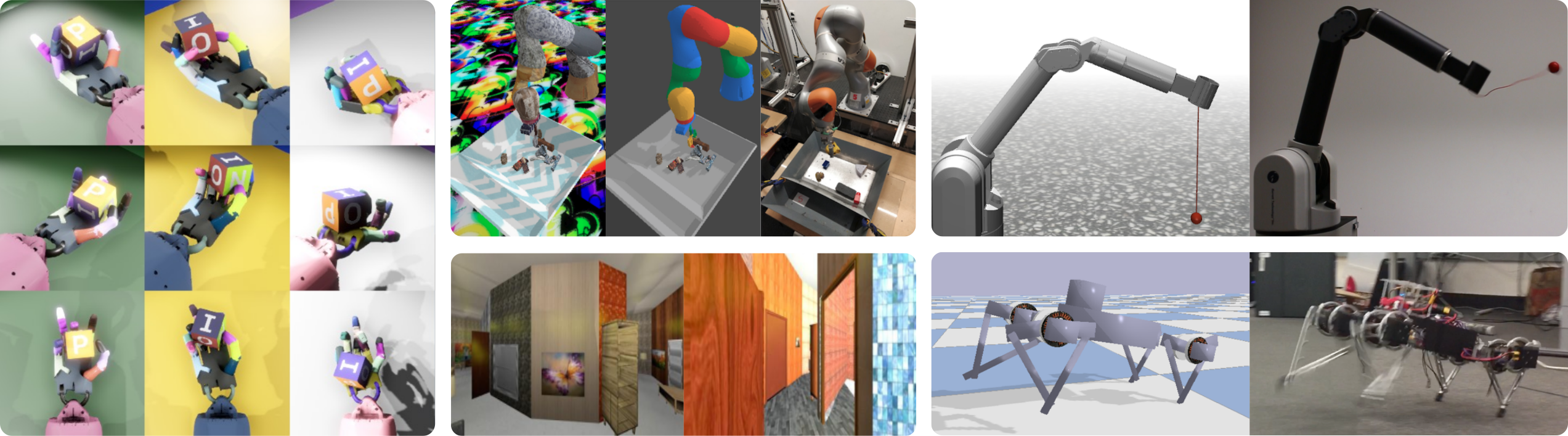}
	\caption[Examples of \simtoreal robot learning research.]{%
		Examples of \simtoreal robot learning research using domain randomization:
		(\textbf{left}) Multiple simulation instances of robotic in-hand manipulation~\citep{OpenAI_18},
		(\textbf{middle top}) transformation to a canonical simulation~\citep{James_Bousmalis_19},
		(\textbf{middle bottom}) synthetic 3D hallways generated for indoor drone flight~\citep{Sadeghi_Levine_17},
		(\textbf{right top}) \bic task solved with adaptive dynamics randomization~\citep{Muratore_Peters_21_RAL},
		(\textbf{right bottom}) quadruped locomotion~\citep{Tan_Vanhoucke_18}.
		\vspace{-0.65\baselineskip}
	}
	\label{fig_sota_collage_sim2real}
\end{figure}

There is a consensus that further increasing the simulator's accuracy alone will not bridge this gap~\citep{Hofner_White_20}.
Looking at breakthroughs in machine learning, we see that deep models in combination with large and diverse data sets lead to better generalization~\citep{Russakovsky_etal_2015,Radford_etal_2019}.
In a similar spirit, a technique called domain randomization has recently gained momentum (Figure~\ref{fig_sota_collage_sim2real}).
The common characteristic of such approaches is the perturbation of simulator parameters, state observations, or applied actions.
Typical quantities to randomize include the bodies' inertia and geometry, the parameters of the friction and contact models, possible delays in the actuation, efficiency coefficients of motors, levels of sensor noise, as well as visual properties such as colors, illumination, position and orientation of a camera, or additional artifacts to the image (e.g., glare).
Domain randomization can be seen as a regularization method that prevents the learner from overfitting to individual simulation instances.
From the Bayesian perspective, we can interpret the distribution over simulators as a representation of uncertainty.

In this paper, we first introduce the necessary nomenclature and mathematical fundamentals for the problem (Section~\ref{sec_sota_mdp}).
Next, we review early approaches for learning from randomized simulations, state the practical requirements, and describe measures for \simtoreal transferability (Section~\ref{sec_sota_simtoreal_foundations}).
Subsequently, we discuss the connections between research on \simtoreal transfer and related fields (Section~\ref{sec_sota_related_fields}).
Moreover, we introduce a taxonomy for domain randomization and categorize the current state of the art (Section~\ref{sec_sota_domain_randomization}).
Finally, we conclude and outline possible future research directions (Section~\ref{sec_sota_discussion_and_outlook}). For those who want to first become more familiar with robot policy learning as well as policy search, we recommend these surveys:~\citet{Kober_Peters_13}, \citet{Deisenroth_Peters_13}, and \citet{Chatzilygeroudis_Mouret_20}.

\section{Problem Formulation and Nomenclature}
\label{sec_sota_mdp}
We begin our discussion by defining critical concepts and nomenclature used throughout this article. 

\noindent\textbf{\acfp{MDP}:} Consider a discrete-time dynamical system
\begin{equation}
	\fs_{t+1} \sim \transprob[\domparams]{\given{\fs_{t+1}}{\fs_t, \fa_t}}, \quad
	\fs_0 \sim \initstatedistr[\domparams]{\fs_0}, \quad
	\fa_t \sim \pol[\polparams]{\given{\fa_t}{\fs_t}}, \quad
	\domparams \sim \domparamdistr{\domparams},
\end{equation}
with the continuous state ${\fs_t \in \stateset[\domparams] \subseteq \RR^{\dimstate}}$ and continuous action ${\fa_t \in \actionset[\domparams] \subseteq \RR^{\dimact}}$ at time step $t$.
The environment, also called domain, is characterized by its parameters ${\domparams \in \RR^{\dimdomparam}}$ (e.g., masses, friction coefficients, time delays, or surface appearance properties) which are in general assumed to be random variables distributed according to an unknown probability distribution ${\domparamdistr{\domparams} \colon \RR^{\dimdomparam} \to \RR^{+}}$.
A special case of this is the common assumption that the domain parameters obey a parametric distribution $\domparamdistr[\domdistrparams]{\domparams}$ with unknown parameters $\domdistrparams$ (e.g., mean and variance). 
The domain parameters determine the transition probability density function ${\transprobsym_{\domparams} \colon \stateset[\domparams] \times \actionset[\domparams] \times \stateset[\domparams] \to \RR^{+}}$ that describes the system's stochastic dynamics.
The initial state $\fs_0$ is drawn from the start state distribution ${\initstatedistrsym_{\domparams} \colon \stateset[\domparams] \to \RR^{+}}$.
In general, the instantaneous reward is a random variable depending on the current state and action as well as the next state.
Here we make the common simplification that the reward is a deterministic function of the the current state and action ${\rewsym_{\domparams} \colon \stateset[\domparams] \times \actionset[\domparams] \to \RR}$.
Together with the temporal discount factor ${\gamma \in [0,1]}$, the system forms a \ac{MDP} described by the tuple ${\mdp[\domparams] = \tuple{\stateset[\domparams], \actionset[\domparams], \transprobsym_{\domparams}, \initstatedistrsym_{\domparams}, \rewsym_{\domparams}, \gamma}}$.

\noindent\textbf{\acf{RL}:} The goal of a \ac{RL} agent is to maximize the expected (discounted) return, a numeric scoring function which measures the policy's performance.
The expected discounted return of a policy $\pol[\polparams]{\given{\fa_t}{\fs_t}}$ with the parameters $\polparams \in \Theta \subseteq \RR^{\dimpolparam}$ is defined as
\begin{equation}
	\label{eq_sota_edr}
	\edr{\polparams,\domparams} =
	\Esub{\fs_0 \sim \initstatedistr*[\domparams]{\fs_0}}{
		\Esub{ \fs_{t+1} \sim \transprob*[\domparams]{\fs_t,\fa_t},\ \fa_t \sim \pol[\polparams]{\given{\fa_t}{\fs_t}} }{ \sum\nolimits_{t=0}^{T-1} \gamma^t \rewfcn[\domparams]{\fs_t, \fa_t} \Big| \polparams, \domparams, \fs_0}
	}.
\end{equation}
While learning from experience, the agent adapts its policy parameters.
The resulting state-action-reward tuples are collected in trajectories, \aka rollouts, $\traj = \set[t=0][T-1]{\fs_t,\fa_t,\rewsym_t} \in \trajspace$ with $\rewsym_t = \rewfcn[\domparams]{\fs_t, \fa_t}$.
In a partially observable \ac{MDP}, the policy's input would not be the state but observations there of ${\fo_t \in \obsset[\domparams] \subseteq \RR^{\dimobs}}$, which are obtained through an environment-specific mapping $\fo_t = f_\text{obs}(\fs_t)$.

\noindent\textbf{Domain randomization:} When augmenting the \ac{RL} setting with domain randomization, the goal becomes to maximize the expected (discounted) return for a distribution of domain parameters
\begin{equation}
	\label{eq_sota_eedr}
	\eedr{\polparams} = 
	\Esub{\domparams \sim \domparamdistr*{\domparams}}{\edr{\polparams,\domparams}} = 
	\Esub{\domparams \sim \domparamdistr*{\domparams}}{
		\Esub{\traj \sim p(\traj)}{\sum\nolimits_{t=0}^{T-1} \gamma^t \rewfcn[\domparams]{\fs_t, \fa_t} \Big| \polparams, \domparams, \fs_0}%
	}.
\end{equation}
The outer expectation with respect to the domain parameter distribution $\domparamdistr{\domparams}$ is the key difference compared to the standard \ac{MDP} formulation.
It enables the learning of robust policies, in the sense that these policies work for a whole set of environments instead of overfitting to a particular problem instance.

\section{Foundations of \SimtoReal Transfer}
\label{sec_sota_simtoreal_foundations}
Modern research on learning from (randomized) physics simulations is based on solid foundation of prior work (Section~\ref{sec_sota_early_methods}).
Parametric simulators are the core component of every \simtoreal method (Section~\ref{sec_sota_simulators}).
Even though the details of their randomization are crucial, they are rarely discussed (Section~\ref{sec_sota_randomizing_simulators}).
Estimating the \simtoreal transferability during or after learning allows one to assess or predict the policy's performance in the target domain (Section~\ref{sec_sota_measureing_sim2real_transferability}).

\subsection{Early Methods}
\label{sec_sota_early_methods}
The roots of randomized simulations trace back to the invention of the Monte Carlo method~\citep{Metropolis_Ulam_49}, which computes its results based on repeated random sampling and subsequent statistical analysis.  
Later, the concept of common random numbers, also called correlated sampling, was developed as a variance reduction technique~\citep{Kahn_Marshall_53,Wright_Ramsay_79}.
The idea is to synchronize the random numbers for all stochastic events across the simulation runs to achieve a (desirably positive) correlation between random variables reducing the variance of an estimator based on a combination of them.
%
Many of the \simtoreal challenges which are currently tackled have already been identified by \citet{Brooks_92}.
In particular, Brooks addresses the overfitting to effects which only occur in simulation as well as the idealized modeling on sensing and actuation.
To avoid overfitting, he advocated for reactive behavior-based programming which is deeply rooted in, hence tailored to, the embodiment.
Focusing on \ac{RL}, \citet{Sutton_91} introduced the Dyna architecture which revolves around predicting from a learned world model and updating the policy from this hypothetical experience.
Viewing the data generated from randomized simulators as \enquote*{imaginary}, emphasizes the parallels of domain randomization to Dyna.
As stated by Sutton, the usage of \enquote*{mental rehearsal} to predict and reason about the effect of actions dates back even further in other fields of research such as psychology~\citep{Craik_43,Dennett_75}.
Instead of querying a learned internal model, \citet{Jakobi_Harvey_95} added random noise the sensors and actuators while learning, achieving the arguably first \simtoreal transfer in robotics.
In follow-up work, \citet{Jakobi_97} formulated the radical envelope of noise hypothesis which states that \enquote{it does not matter how inaccurate or incomplete [the simulations] are: controllers that have evolved to be reliably fit in simulation will still transfer into reality.}
Picking up on the idea of common random numbers, \citet{Ng_Jordan_00} suggested to explicitly control the randomness of a simulator, i.e., the random number generator's state, rendering the simulator deterministic.
This way the same initial configurations can be (re-)used for Monte Carlo estimations of different policies' value functions, allowing one to conduct policy search in partially observable problems.
\citet{Bongard_etal_2006} bridged the \simtoreal gap through iterating model generation and selection depending on the short-term state-action history.
This process is repeated for a given number of iterations, and then yields the self-model, i.e., a simulator, which best explains the observed data.

Inspired by these early approaches, the systematic analysis of randomized simulations for robot learning has become a highly active research direction.  
Moreover, the prior work above also falsifies the common belief that domain randomization originated recently with the rise of deep learning.
Nevertheless, the current popularity of domain randomization can be explained by its widespread use in the computer vision and locomotion communities as well as its synergies with deep learning methods.
The key difference between the early and the recent domain randomization methods (Section~\ref{sec_sota_domain_randomization}) is that the latter (directly) manipulate the simulators' parameters.

\subsection{Constructing Stochastic Simulators}
\label{sec_sota_simulators}
Simulators can be obtained by implementing a set of physical laws for a particular system.
Given the challenges in implementing an efficient simulator for complex systems, it is common to use general purpose physics engines such as
\href{https://ode.org}{ODE},
\href{https://dartsim.github.io/}{DART},
\href{https://pybullet.org/wordpress/}{Bullet},
\href{http://newtondynamics.com/forum/newton.php}{Newton},
\href{https://simtk.org/projects/simbody}{SimBody},
\href{https://www.cm-labs.com/vortex-studio/}{Vortex},
\href{http://www.mujoco.org/}{MuJoCo},
\href{https://www.havok.com/}{Havok},
\href{https://projectchrono.org/}{Chrono},
\href{https://raisim.com/}{RaiSim},
\href{https://developer.nvidia.com/physx-sdk}{PhysX},
\href{https://developer.nvidia.com/flex}{FleX}, or
\href{https://github.com/google/brax}{Brax}.
These simulators are parameterized generative models, which describe how multiple bodies or particles evolve over time by interacting with each other.
The associated physics parameters can be estimated by system identification (Section~\ref{sec_sota_system_identification}), which generally involves executing experiments on the physical platform and recording associated measurement. 
Additionally, using the Gauss-Markov theorem one could also compute the parameters' covariance and hence construct a normal distribution for each domain parameter. 
Differentiable simulators facilitate deep learning for robotics~\citep{Degrave_Wyffels_19,Coumans_TinyDiffSim,Heiden_Millard_21} by propagating the gradients though the dynamics.
Current research extends the differentiability to soft body dynamics~\citep{Hu_Matusik_19}.
Alternatively, the system dynamics can be captured using nonparametric methods like \acp{GP}~\citep{Rasmussen_Williams_2006} as for example demonstrated by \citet{Calandra_Peters_15}.
It is important to keep in mind that even if the domain parameters have been identified very accurately, simulators are nevertheless just approximations of the real world and are thus always imperfect.

Several comparisons between various physics engines were made~\citep{Ivaldi_Nori_14,Erez_Todorov_15,Chung_Pollard_16,Collins_Leitner_19,Korber_Gluck_21}.
However, note that these results become outdated quickly due to the rapid development in the field, or are often limited to very few scenarios and partially introduce custom metrics to measure their performance or accuracy.

Apart from the physics engines listed above, there is an orthogonal research direction investigating human-inspired learning of the physics laws from visual input~\citep{Battaglia_Tenenbaum_13,Wu_Tenenbaum_15} as well as physical reasoning given a configuration of bodies~\citep{Battaglia_Kavukcuoglu_16}, which is out of the scope of this review. 

\subsection{Randomizing a Simulator}
\label{sec_sota_randomizing_simulators}
Learning from randomized simulations entails significant design decisions:

\textbf{Which parameters should be randomized?}
Depending on the problem, some domain parameters have no influence (e.g., the mass of an idealized rolling ball) while others are pivotal (e.g., the pendulum length for a stabilization task).
It is recommended to first identify the essential parameters~\citep{Xie_Garg_20}.
For example, most robot locomotion papers highlight the importance of varying the terrain and contact models, while applications such as drone control benefit from adding perturbations, e.g., to simulate a gust of wind.
Injecting random latency and noise to the actuation is another frequent modeling choice.
Starting from a small set of randomized domain parameters, identified from prior knowledge, has the additional benefit of shortening the evaluation time which involves approximating an expectation over domains, which scales exponentially with the number of parameters.
Moreover, including at least one visually observable parameter (e.g., an extent of a body) helps to verify if the values are set as expected.

\textbf{When should the parameters be randomized?}
Episodic dynamics randomization, without a rigorous theoretical justification, is the most common approach.
Randomizing the domain parameters at every time step instead would drastically increase the variance, and pose a challenge to the implementations since this typically implies recreating the simulation at every step.
Imagine a stack of cubes standing on the ground.
If we now vary the cubes' side lengths individually while keeping their absolute positions fixed, they will either lose contact or intersect with their neighboring cube(s).
In order to keep the stack intact, we need to randomize the cubes with respect to their neighbors, additionally moving them in space.
Executing this once at the beginning is fine, but doing this at every step creates artificial \enquote{movement} which would almost certainly be detrimental.
Orthogonal to the argumentation above, alternative approaches apply random disturbance forces and torques at every time step.
In these cases, the distribution over disturbance magnitudes is chosen to be constant until the randomization scheme is updated. 
To the best of our knowledge, event-triggered randomization has not been explored yet.

\textbf{How should the parameters be randomized?}
Answering this question is what characterizes a domain randomization method (Section~\ref{sec_sota_domain_randomization}).
There are a few aspects that needs to be considered in practice when designing a domain randomization scheme, such as the numerical stability of the simulation instances.
Low masses for example quickly lead to stiff differential equations which might require a different (implicit) integrator.
Furthermore, the noise level of the introduced randomness needs to match the precision of the state estimation.
If the noise is too low, the randomization is pointless.
On the other side, if the noise level is too high, the learning procedure will fail.
To find the right balance between these considerations, we can start by statistically analyzing the incoming measurement signals.

\textbf{What about physical plausibility?}
The application of pseudo-random color patterns, e.g., Perlin noise~\citep{Perlin_02}, has become a frequent choice for computer vision applications.
Despite that these patterns do not occur on real-world objects, this technique has improved the robustness of object detectors~\citep{James_Johns_17,Pinto_Abbeel_17}.
Regarding the randomization of dynamics parameters, no research has so far hinted that physically implausible simulations (e.g., containing bodies with negative masses) are useful.
On the other hand, it is safe to say that these can cause numerical instabilities.
Thus, ensuring feasibility of the resulting simulator is highly desirable.
One solution is to project the domain parameters into a different space, guaranteeing physical plausibility via the inverse projection.
For example, a body's mass could be learned in the log-space such that the subsequent exp-transformation, applied before setting the new parameter value, yields strictly positive numbers.
However, most of the existing domain randomization approaches can not guarantee physical plausibility.

Even in the case of rigid body dynamics there are notable differences between physics engines, as was observed by \citet{Muratore_Peters_18_CoRL} when transferring a robot control policy trained using Vortex to Bullet and vice versa.
Typical sources for deviations are different coordinate representations, numerical solvers, friction and contact models.
Especially the latter two are decisive for robot manipulation.
For vision-based tasks, \citet{Alghonaim_Johns_20} found a strong correlation between the renderer's quality and \simtoreal transferability.
Additionally, the authors emphasize the importance of randomizing both distractor objects and background textures for generalizing to unseen environments.

\subsection{Measuring and Predicting the Reality Gap}
\label{sec_sota_measureing_sim2real_transferability}
Coining the term \enquote*{reality gap}, \citet{Koos_Doncieux_10} hypothesize that the fittest solutions in simulation often rely on poorly simulated phenomena.
From this, they derive a multi-objective
formulation for \simtoreal transfer where performance and transferability need to be balanced.
In subsequent work, \citet{Koos_Doncieux_13} defined a transferability function that maps controller parameters to their estimated target domain performance.
A surrogate model of this function is regressed from the real-world fitness values that are obtained by executing the controllers found in simulation.

The \ac{SOB}~\citep{Muratore_Peters_18_CoRL,Muratore_Peters_21_PAMI} is a quantitative measure for the transferability of a control policy from a set of source domains to a different target domain originating from the same distribution.
Building on the formulation of the optimality gap from convex optimization~\citep{Mak_etal_1999,Bayraksan_Morton_2006}, \citet{Muratore_Peters_18_CoRL} proposed a Monte Carlo estimator of the \ac{SOB} as well as an upper confidence bound, tailored to reinforcement learning settings.
This bound can be used as an indicator to stop training when the predicted transferability exceeds a threshold.
Moreover, the authors show that the \ac{SOB} is always positive, i.e. optimistic, and in expectation monotonically decreases with an increasing number of domains. 

\citet{Collins_Leitner_19} quantify the accuracy of ODE, (Py)Bullet, Newton, Vortex, and MuJoCo in a real-world robotic setup.
The accuracy is defined as the accumulated mean-squared error between the Cartesian ground truth position, tracked by a motion capture system, and the simulators' prediction.
Based on this measure, they conclude that simulators are able to model the control and kinematics accurately, but show deficits during dynamic robot-object interactions.

To obtain a quantitative estimate of the transferability, \citet{Zhang_Zaremba_20} suggest to learn a probabilistic dynamics model which is evaluated on a static set of target domain trajectories.
This dynamics model is trained jointly with the policy in the same randomized simulator.
The transferability score is chosen to be the average negative log-likelihood of the model's output given temporal state differences from the real-world trajectories.
Thus, the proposed method requires a set of pre-recorded target domain trajectories, and makes the assumption that for a given domain the model's prediction accuracy correlates with the policy performance.

With robot navigation in mind, \citet{Kadian_Truong_Batra_20} define the \ac{SRCC} to be the Pearson correlation coefficient on data pairs of scalar performance metrics.
The data pairs consist of the policy performance achieved in a simulator instance as well as in the real counterpart.
Therefore, in contrast to the \ac{SOB}~\citep{Muratore_Peters_18_CoRL}, the \ac{SRCC} requires real-world rollouts.
A high \ac{SRCC} value, i.e. close to 1, predicts good transferability, while low values, i.e. close to 0, indicates that the agent is exploited the simulation during learning.
\citet{Kadian_Truong_Batra_20} also report tuning the domain parameters with grid search to increase the \ac{SRCC}.
By using the Pearson correlation, the \ac{SRCC} is restricted to linear correlation, which might not be a notable restriction in practice.

\section{Relation of \SimtoReal to other Fields}
\label{sec_sota_related_fields}
\begin{figure}[t]
	\centering
	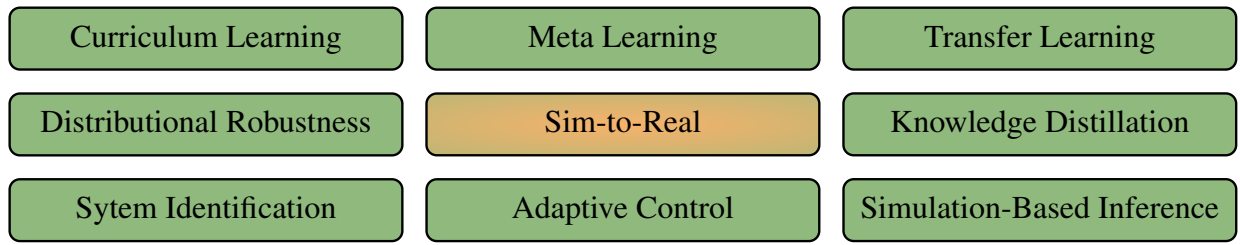
	\caption[Overview of the \simtoreal research and related fields.]{
		Topological overview of the \simtoreal research and a selection of related fields.
	}
	\label{fig_sota_realated fields}
\end{figure}
There are several research areas that overlap with \simtoreal in robot learning, more specifically domain randomization (Figure~\ref{fig_sota_realated fields}).
In the following, we describe those that either share the same goal, or employ conceptually similar methods.

\subsection{Curriculum Learning}
\label{sec_sota_curriculum_learning}
The key idea behind curriculum learning is to increase the sample efficiency by scheduling the training process such that the agent first encounters \enquote*{easier} tasks and gradually progresses to \enquote*{harder} ones.
Hence, the agent can bootstrap from the knowledge it gained at the beginning, before learning to solve more difficult task instances.
Widely known in supervised learning~\citep{Bengio_Weston_09,Kumar_Koller_10}, curriculum learning has been applied to \ac{RL}~\citep{Asada_Hosoda_96,Erez_Smart_08,Klink_Peters_19,Klink_Pajarinen_21}.
The connection between curriculum learning and domain randomization can be highlighted by viewing the task as a part of the domain, i.e., the \ac{MDP}, rendering the task parameters a subspace of the domain parameters.
From this point of view, the curriculum learning schedule describes how the domain parameter distribution is updated. 
There are several challenges to using a curriculum learning approach for \simtoreal transfer.
Three such challenges are: (i) we can not always assume to have an assessment of the difficulty level of individual domain parameter configurations, (ii) curriculum learning does not aim at finding solutions robust to model uncertainty, and (iii) curriculum learning methods may require a target distribution which is not defined in the domain randomization setting.
However, adjustments can be made to circumvent these problems. 
\citet{OpenAI_19} suggested a heuristic for the domain randomization schedule that increases the boundaries of each domain parameter individually until the return drops more than a predefined threshold.
Executing this approach on a computing cluster, the authors managed to train a policy and a vision system which in combination solve a Rubik's cube with a tendon-driven robotic hand.
Another intersection point of curriculum learning and \simtoreal transfer is the work by \citet{Morere_Ramos_19}, where a hierarchical planning method for discrete domains with unknown dynamics is proposed.
Learning abstract skills based on a curriculum enables the algorithm to outperform planning and \ac{RL} baselines, even in domains with a very large number of possible states.

\subsection{Meta Learning}
\label{sec_sota_meta_learning}
Inspired by the human ability to quickly master new tasks by leveraging the knowledge extracted from solving other tasks, meta learning~\citep{Santoro_Lillicrap_16,Finn_Levine_17} seeks to make use of prior experiences gained from conceptually similar tasks.
The field of meta learning currently enjoys high popularity, leading to abundant follow-up work.
\citet{Grant_Griffiths_18} for example casts meta learning as hierarchical Bayesian inference.
Furthermore, the meta learning framework has been adapted to the \ac{RL} setting~\citep{Wang_Botvinick_17,Nagabandi_Clavera_Finn_19}.
The optimization over an ensemble of tasks can be translated to the optimization over an ensemble of domain instances, modeled by different \acp{MDP} (Section~\ref{sec_sota_mdp}).
Via this duality one can view domain randomization as a special form of meta learning where the robot's task remains qualitatively unchanged but the environment varies.
Thus, the tasks seen during the meta training phase are analogous to domain instances experienced earlier in the training process.
However, when looking at the complete procedure, meta learning and domain randomization are fundamentally different.
The goal of meta learning, i.e., \citet{Finn_Levine_17}, is to find a suitable set of initial weights, which when updated generalizes well to a new task.
Domain randomization on the other hand strives to directly solve a single task, generalizing over domain instances.


\subsection{Transfer Learning}
\label{sec_sota_transfer_learning}
The term transfer learning covers a wide range of machine learning research, aiming at using knowledge learned in the source domain to solve a task in the target domain.
Rooted in classification, transfer learning is categorized in several subfields by for example differentiating (i) if labeled data is available in the source or target domain, and (ii) if the tasks in both domains are the same~\citep{Pan_Yang_10,Zhuang_He_21}.
Domain adaptation is one of the resulting subfields, specifying the case where ground truth information is only available in the target domain which is not equal to the source domain while the task remains the same.
Thus, domain adaptation methods are in general suitable to tackle \simtoreal problems.
However, the research fields evolved at different times in different communities, with different goals in mind.
The keyword \enquote*{sim-to-real} specifically concerns regression and control problems where the focus lies on overcoming the mismatch between simulation and reality.
In contrast, most domain adaptation techniques are not designed for a dynamical system as the target domain.

\subsection{Knowledge Distillation}
\label{sec_sota_knowledge_distillation}
When executing a controller on a physical device operating at high frequencies, it is of utmost importance that the forward pass finishes with the given time frame.
With deep \ac{NN} policies, and especially with ensembles of these, this requirement can become challenging to meet.
Distilling the knowledge of a larger network into a smaller one reduces the evaluation time.
Knowledge distillation~\citep{Hinton_Dean_15} has been successfully applied to several machine learning applications such as natural language processing~\citep{Cui_Rosenberg_17}, and object detection~\citep{Chen_Chandraker_17}.
In the context of \ac{RL}, knowledge distillation techniques can be used to compress the learned behavior of one or more teachers into a single student~\citep{Rusu_Hadsell_16_ICLR}.
Based on samples generated by the teachers, the student is trained in a supervised manner to imitate them.
This idea can be applied to \simtoreal robot learning in a straightforward manner, where the teachers can be policies optimal for specific domain instances~\citep{Brosseit_Hahner_Muratore_Peters_21}.
Complementarily, knowledge distillation has been applied to multitask learning~\citep{Parisotto_Salakhutdinov_15,Teh_Pascanu_17}, promising to improve sample efficiency when learning a new task.
A technical comparison of policy distillation methods for \ac{RL} is provided by \citet{Czarnecki_Jaderberg_19}.

\subsection{Distributional Robustness}
\label{sec_sota_distributional_robustness}
The term robustness is overloaded with different meanings, such as the ability to (quickly) counteract external disturbances, or the resilience against uncertainties in the underlying model's parameters.
The field of robust control aims at designing controllers that explicitly deal with these uncertainties~\citep{Zhou_Doyle_98}.
Within this field, distributional robust optimization is a framework to find the worst-case probabilistic model from a so-called ambiguity set, and subsequently set a policy which acts optimally in this worst case.
Mathematically, the problem is formulated as bilevel optimization, which is solved iteratively in practice.
By restricting the model selection to the ambiguity set, distributional robust optimization regularizes the adversary to prevent the process from yielding solutions that are overly conservative policies.
Under the lens of domain randomization, the ambiguity set closely relates to the distribution over domain parameters.
\citet{Abdulsamad_Dorau_Peters_21} for example define the ambiguity set as a \ac{KL} ball the nominal distribution.
Other approaches use a moment-based ambiguity set~\citep{Delage_Ye_10} or introduce chance constrains~\citep{VanPary_Morari_16}.
For a review of distributional robust optimization, see \citet{Zhen_Wiesemann_21}.
\cite{Chatzilygeroudis_Mouret_20} point out that performing policy search under an uncertain model is equivalent to finding a policy that can perform well under various dynamics models.
Hence, they argue that \enquote{model-based policy search with probabilistic models is performing something similar to dynamics randomization}.

\subsection{System Identification}
\label{sec_sota_system_identification}
The goal of system identification is to find the set of model parameters which fit the observed data best, typically by minimizing the prediction-dependent loss such as the mean-squared error.
Since the simulator is the pivotal element in every domain randomization method, the accessible parameters and their nominal values are of critical importance.
When a manufacturer does not provide data for all model parameters, or when an engineer wants to deploy a new model, system identification is typically the first measure to obtain an estimate of the domain parameters.
%
In principle, a number of approaches can be applied depending on the assumptions on the internal structure of the simulator.
The earliest approaches in robotics recognized the linearity of the rigid body dynamics with respect to combinations of physics parameters such as masses, moments of inertia, and link lengths, thus proposed to use linear regression~\citep{Atkeson_Hollerbach_86}, and later Bayesian linear regression~\citep{Ting_Nakanishi_06}.
However, it was quickly observed that the inferred parameters may be physically implausible, leading to the development of methods that can account for this~\citep{Ting_Schaal_11}.
With the advent of deep learning, such structured physics-based approaches have been enhanced with \acp{NN}, yielding nonlinear system identification methods such as the ones based on the Newton-Euler forward dynamics~\citep{Sutanto_Meier_20,Lutter_Silberbauer_Peters_21}.
Alternatively, the simulator can be augmented with a \ac{NN} to learn the domain parameter residuals, minimizing the one step prediction error~\citep{Allevato_Thomaz_19}.
%
On another front, system identification based on the classification loss between simulated and real samples has been investigated~\citep{Jiang_Tan_21,Du_Watkins_Pathak_21}.
System identification can also be interpreted as an episodic \ac{RL} problem by treating the trajectory mismatch as the cost function and iteratively updating a distribution over models~\citep{Chebotar_Fox_19}.
Recent simulation-based inference methods yield highly expressive posterior distributions that capture multi-modality as well as correlations between the domain parameters (Section~\ref{sec_sota_sbi}).

\subsection{Adaptive Control}
\label{sec_sota_adaptive_control}
The well-established field of adaptive control is concerned with the problem of adapting a controller's parameters at runtime to operate initially uncertain or varying systems (e.g., aircraft reaching supersonic speed).
A prominent method is model reference adaptive control, which tracks a reference model's output specifying the desired closed-loop behavior.
\ac{MIAC} is a different variant, which includes an online system identification component that continuously estimates the system's parameters based on the prediction error of the output signal~\citep{Astrom_Wittenmark_08,Landau_Karimi_11}.
Given the identified system, the controller is updated subsequently.
Similarly, there exists a line of \simtoreal reinforcement learning approaches that condition the policy on the estimated domain parameters~\citep{Yu_Turk_17,Yu_Turk_19,Mozifian_Dudek_20} or a latent representation thereof~\citep{Yu_Liu_19,Peng_Levine_20,Kumar_Malik_21}.
The main difference to \ac{MIAC} lies in the adaption mechanism.
Adaptive control techniques typically define the parameters' gradient  proportional to the prediction error, while the approaches referenced above make the domain parameters an input to the policy.

\subsection{Simulation-Based Inference}
\label{sec_sota_sbi}
Simulators are predominantly used as forward models, i.e., to make predictions.
However, with the increasing fidelity and expressiveness of simulators, there is a growing interest to also use them for probabilistic inference~\citep{Cranmer_Louppe_19}.
In the case of simulation-based inference, the simulator and its parameters define the statistical model.
Inference tasks differ by the quantity to be inferred.
Regarding \simtoreal transfer, the most frequent task is to infer the simulation parameters from real-world time series data.
Similarly to system identification (Section~\ref{sec_sota_system_identification}), the result can be a point estimate, or a posterior distribution.
\ac{LFI} methods are a type of simulation-based inference approaches which are particularly well-suited when we can make very little assumptions about the underlying generative model, treating it as an implicit function.
These approaches only require samples from the model (e.g., a non-differentiable black-box simulator) and a measure of how likely real observations could have been generated from the simulator.
Approximate Bayesian computation is well-known class of \ac{LFI} methods that applies Monte Carlo sampling to infer the parameters by comparing summary statistics of synthetically generated and observed data.
There exist plenty of variants for approximate Bayesian computation~\citep{Marjoram_Tavare_03,Beaumont_Robert_09,Sunnaker_Dessimoz_13} as well as studies on the design of low-dimensional summary statistics~\citep{Fearnhead_Prangle_12}.
In order to increase the efficiency and thereby scale \ac{LFI} higher-dimensional problems, researchers investigated amortized approaches, which conduct the inference over multiple sequential rounds.
Sequential neural posterior estimation approaches~\citep{Papamakarios_Murray_16,Lueckmann_Macke_17, Greenberg_Macke_19} approximate the conditional posterior, allowing for direct sampling from the posterior.
Learning the likelihood~\citep{Papamakarios_Murray_19} can be useful in the context for hypothesis testing.
Alternatively, posterior samples can be generated from likelihood-ratios~\citep{Hermans_Louppe_20,Durkan_Papamakarios_20}.
However, simulation-based inference does not explicitly consider policy optimization or domain randomization.
Recent approaches connected all three techniques, and closed the reality gap by inferring a distribution over simulators while training policies in simulation~\citep{Ramos_Fox_19,Barcelos_Ramos_20,Muratore_Peters_21_CoRL}.

\section{Domain Randomization for \SimtoReal Transfer}
\label{sec_sota_domain_randomization}
\begin{wrapfigure}[7]{r}{0.51\columnwidth} 
	\vspace*{-1.2\baselineskip}
	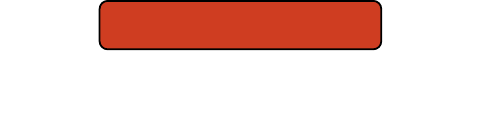
	\caption[Topological overview of domain randomization]{Topological overview of domain randomization methods.}
	\label{fig_sota_domain_randomization}
\end{wrapfigure}
We distinguish between static (Section~\ref{sec_sota_static_dr}), adaptive (Section~\ref{sec_sota_adaptive_dr}), and adversarial (Section~\ref{sec_sota_adversarial_dr}) domain randomization.
Static, as well as adaptive, methods are characterized by randomly sampling a set of domain parameters $\domparams \sim \domparamdistr{\domparams}$ at the beginning of each simulated rollout.
A randomization scheme is categorized as adaptive if the domain parameter distribution is updated during learning, otherwise the scheme is called static.
The main advantage of adaptive schemes is that they alleviate the need for hand-tuning the distributions of the domain parameters, which is currently a decisive part of the hyper-parameter search in a static scheme.
Nonetheless, the prior distributions still demand design decisions.
On the downside, every form of adaptation requires data from the target domain, typically the real robot, which is significantly more expensive to obtain.
Another approach for learning robust policies in simulation is to apply adversarial disturbances during the training process. 
We classify these perturbations as a form of domain randomization, since they either depend on a highly stochastic adversary learned jointly with the policy, or directly contain a random process controlling the application of the perturbation.
Adversarial approaches may yield exceptionally robust control strategies.
However, without any further restrictions, it is always possible to create scenarios in which the protagonist agent can never win, i.e., the policy can not learn the task.
Balancing the adversary's power is pivotal to an adversarial domain randomization method, adding a sensitive hyper-parameter.

Another way to distinguish domain randomization concepts is the representation of the domain parameter distribution.
The vast majority of algorithms assume a specific probability distribution (e.g., normal or uniform) independently for every parameter.
This modeling decision has the benefit of greatly reducing the complexity, but at the same time severely limits the expressiveness.
Novel \acs{LFI} methods (Section~\ref{sec_sota_adaptive_dr}) estimate the complete posterior, hence allow the recognition of correlations between the domain parameters, multi-modality, and skewness.

\subsection{Static Domain Randomization}
\label{sec_sota_static_dr}
\begin{wrapfigure}[8]{r}{0.50\columnwidth} 
	\raggedleft
	\vspace*{-1.35\baselineskip}
	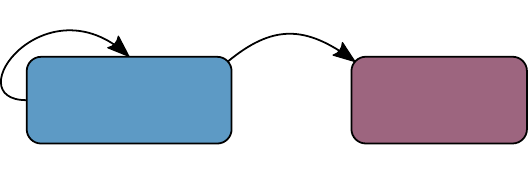
	\vspace*{-2.3\baselineskip}
	\caption{Conceptual illustration of static domain randomization.}
	\label{fig_sota_static_dr}
\end{wrapfigure}
Approaches that sample from a fixed domain parameter distribution typically aim at performing \simtoreal transfer without using any real-world data.
Since running the policy on a physical device is generally the most difficult and time-consuming part, static approaches promise quick and relatively easy to obtain results.
In terms of final policy performance in the target domain, these methods are usually inferior to those that adapt the domain parameter distribution.
Nevertheless, static domain randomization has bridged the reality gap in several cases.

\paragraph*{Randomizing Dynamics without Using Real-World Data at Runtime}
More than a decade ago, \citet{Wang_Hertzmann_10} proposed to randomize the simulator in which the training data is generated.
The authors examined the randomization of initial states, external disturbances, goals, and actuator noise, clearly showing an improved robustness of the learned locomotion controllers in simulated experiments (\simtosim).
\citet{Mordatch_Todorov_15} used a finite model ensembles to run (offline) trajectory optimization on a small-scale humanoid robot, achieving one of the first \simtoreal transfers in robotics powered by domain randomization.
Similarly,~\citet{Lowrey_etal_2018} employed the \acl{NPG}~\citep{Kakade_2001} to learn a continuous controller for a three-finger positioning task, after carefully identifying the system's parameters.
Conforming with~\citet{Mordatch_Todorov_15}, their results showed that the policy learned from the identified model was able to perform the \simtoreal transfer, but the policies learned from an ensemble of models was more robust to modeling errors.
In contrast, \citet{Peng_Abbeel_18} combined model-free \ac{RL} with recurrent \ac{NN} policies that were trained using hindsight experience replay~\citep{Andrychowicz_Zaremba_17} in order to push an object by controlling a robotic arm.
\citet{Tan_Vanhoucke_18} presented an example for learning quadruped gaits from randomized simulations, where particular efforts were made to conduct a prior system identification.
They empirically found that sampling domain parameters from a uniform distribution together with applying random forces and regularizing the observation space can be enough to cross the reality gap.
For quadrotor control, \citet{Molchanov_Chen_Sukhatme_19} trained feedforward \ac{NN} policies which generalize over different physical drones.
The suggested randomization includes a custom model for motor lag and noise based on an Ornstein-Uhlenbeck process.
\citet{Rajeswaran_Levine_16} explored the use of a risk-averse objective function, optimizing a lower quantile of the return.
The method was only evaluated on simulated MuJoCo tasks, however it was also one of the first methods that draws upon the Bayesian perspective.
Moreover, this approach was employed as a baseline by~\citet{Muratore_Peters_21_PAMI}, who introduced a measure for the inter-domain transferability of controllers together with a risk-neutral randomization scheme.
The resulting policies have the unique feature of providing a (probabilistic) guarantee on the estimated transferability and managed to directly transfer to the real platform in two different experiments.
\cite{Siekmann_Hurst_21} achieved the \simtoreal transfer of a recurrent \ac{NN} policy for bipedal walking.
The policy was trained using model-free \ac{RL} in simulation with uniformly distributed dynamics parameters as well as randomized task-specific terrain.
According to the authors, the recurrent architecture and the terrain randomization were pivotal.

\paragraph*{Randomizing Dynamics Using Real-World Data at Runtime}
The work by~\citet{Cully_etal_2015} can be seen as both static and adaptive domain randomization, where a large set of hexapod locomotion policies is learned before execution on the physical robot, and subsequently evaluated in simulation.
Every policy is associated with one configuration of the so-called behavioral descriptors, which can be interpreted as domain parameters.
Instead of retraining or fine-tuning, the proposed algorithm reacts to performance drops, e.g., due to damage, by querying \ac{BO} to sequentially select one of the pretrained policies and measure its performance on the robot.
Instead of randomizing the simulator parameters, \cite{Cutler_How_15} explored learning a probabilistic model, chosen to be a \ac{GP}, of the environment using data from both simulated and real-world dynamics.
A key feature of this method is to incorporate the simulator as a prior for the probabilistic model, and subsequently use this information of the policy updates with PILCO~\citep{Deisenroth_Rasmussen_11}.
The authors demonstrated policy transfer for a inverted pendulum task.
In follow-up work, \citet{Cutler_How_16} extended the algorithm to make a remote-controlled toy car learn how to drift in circles.
\citet{Antonova_Rai_Kragic_19} propose a sequential \ac{VAE} to embed trajectories into a compressed latent space which is used with \ac{BO} to search for controllers.
The \ac{VAE} and the domain-specific high-level controllers are learned jointly, while the randomization scheme is left unchanged.
Leveraging a custom kernel which measures the \ac{KL} divergence between trajectories and the data efficiency of \ac{BO}, the authors report successful \simtoreal transfers after 10 target domain trials for a hexapod locomotion task as well as 20 trials for a manipulation task.
\citet{Kumar_Malik_21} learned a quadruped locomotion policy that passed joint positions to a lower level \acs{PD} controller without using any real-wold data.
The essential components of this approach are the encoder that projects the domain parameters to a latent space and the adaption module which is trained to regress the latent state from the recent history of measured states and actions.
The policy is conditioned on the current state, the previous actions, and the latent state which needs to be reconstructed during deployment in the physical world.
Emphasizing the importance of the carefully engineered reward function, the authors demonstrate the method's ability to transfer from simulation to various outdoor terrains.

\paragraph*{Randomizing Visual Appearance and Configurations}
\citet{Tobin_Abbeel_17} learned an object detector for robot grasping using a fixed domain parameter distribution, and bridged the gap with a deep \ac{NN} policy trained exclusively on simulated \acs{RGB} images.
Similarly, \citet{James_Johns_17} added various distracting shapes as well as structured noise~\citep{Perlin_02} when learning a robot manipulation task with an end-to-end controller that mapped pixels to motor velocities. 
The approach presented by \citet{Pinto_Abbeel_17} combines the concepts of static domain randomization and actor-critic training~\citep{Lillicrap_Wierstra_15}, enabling the direct \simtoreal transfer of the abilities to pick, push, or move objects.
While the critic has access to the simulator's full state, the policy only receives images of the environment, creating an information asymmetry.
\citet{Matas_Davison_18} used the asymmetric actor-critic idea from \citet{Pinto_Abbeel_17} as well as several other improvements to train a deep \ac{NN} policy end-to-end, seeded with prior demonstrations.
Solving three variations of a tissue folding task, this work scales \simtoreal visuomotor manipulation to deformable objects.
Purely visual domain randomization has also been applied to aerial robotics, where \citet{Sadeghi_Levine_17} achieved \simtoreal transfer for learning to fly a drone through indoor environments.
The resulting deep \ac{NN} policy was able to map from monocular images to normalized 3D drone velocities.
Similarly, \citet{Polvara_Patacchiola_Neumann_20} demonstrated landing of a quadrotor trained in end-to-end fashion using randomized environments.
\citet{Dai_Arulkumaran_Bharath_19} investigated the effect of domain randomization on visuomotor policies, and observed that this leads to more redundant and entangled representations accompanied with significant statistical changes in the weights.
\citet{Yan_Pinto_20} apply \ac{MPC} to manipulate of deformable objects using a forward model based on visual input.
The novelty of this approach is that the predictive model is trained jointly with an embedding to minimizing a contrastive loss~\citep{Oord_Vinyals_18} in the latent space.
Finally, domain randomization was applied to transfer the behavior from simulation to the real robot.

\paragraph*{Randomizing Dynamics, Randomizing Visual Appearance, and Configurations}
Combining \acp{GAN} and domain randomization, \citet{Bousmalis_Irpan_Wohlhart_Vanhoucke_18} greatly reduced the number of necessary real-world samples for learning a robotic grasping task.
The essence of their method is to transform simulated monocular \acs{RGB} images in a way that is closely matched to the real counterpart.
Extensive evaluation on the physical robot showed that domain randomization as well as the suggested pixel-level domain adaptation technique were important to successfully transfer.
Despite the pixel-level domain adaptation technique being learned, the policy optimization in simulation is done with a fixed randomization scheme.
In related work \citet{James_Bousmalis_19} train a \ac{GAN} to transform randomized images to so-called canonical images, such that a corresponding real image would be transformed to the same one.
This approach allowed them to train purely from simulated images, and optionally fine-tune the policy on target domain data.
Notably, the robotic in-hand manipulation conducted by~\citet{OpenAI_18} demonstrated that domain randomization in combination with careful model engineering and the usage of recurrent \acp{NN} enables \simtoreal transfer on an unprecedentedly difficulty level.


\subsection{Adaptive Domain Randomization}
\label{sec_sota_adaptive_dr}
\begin{wrapfigure}[10]{r}{0.50\columnwidth} 
	\raggedleft
	\vspace*{-1.3\baselineskip}
	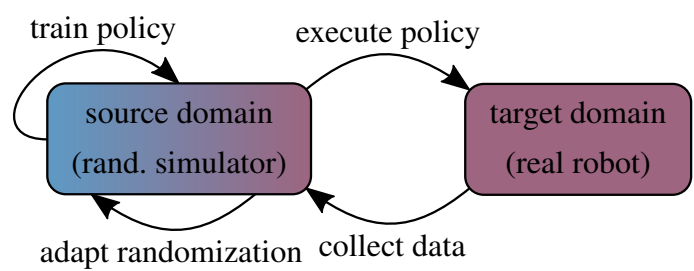
	\caption{Conceptual illustration of adaptive domain randomization.}
	\label{fig_sota_adaptive_dr}
\end{wrapfigure}
Static domain randomization (Section~\ref{sec_sota_static_dr}) is inherently limited and implicitly assumes knowledge of the true mean of the domain parameters or accepts biased samples.
Adapting the randomization scheme allows the training to narrow or widen the search distribution in order to fulfill one or multiple criteria which can be chosen freely.
The mechanism devised for updating the domain parameter distribution as well as the procedure to collect meaningful target domain data are typically the center piece of adaptive randomization algorithms.
In this process the execution of intermediate policies on the physical device is the most likely point of failure.
However, approaches that update the distribution solely based on data from the source domain are less flexible and generally less effective.

\paragraph*{Conditioning Policies on the Estimated Domain Parameters}
\citet{Yu_Turk_17} suggested the use of a \ac{NN} policy that is conditioned on the state and the domain parameters.
Since these parameters are not assumed to be known, they have to be estimated, e.g., with online system identification.
For this purpose, a second \ac{NN} is trained to regress the domain parameters from the observed rollouts. 
By applying this approach to simulated continuous control tasks, the authors showed that adding the online system identification module can enable an adaption to sudden changes in the environment.
In subsequent research, \citet{Yu_Liu_19} intertwined policy optimization, system identification, and domain randomization.
The proposed method first identifies bounds on the domain parameters which are later used for learning from the randomized simulator.
In a departure from their previous approach, the policy is conditioned on a latent space projection of the domain parameters.
After training in simulation, a second system identification step runs \ac{BO} for a fixed number of iterations to find the most promising projected domain parameters.
The algorithm was evaluated on \simtoreal bipedal robot walking.
\citet{Mozifian_Dudek_20} also introduce a dependence of the policy \wrt to the domain parameters.
These are updated by gradient ascent on the average return over domains, regularized by a penalty proportional to the \ac{KL} divergence.
Similar to \citet{Ruiz_Chandraker_19}, the authors update the domain parameter distribution using the score function gradient estimator.
\citet{Mozifian_Dudek_20} tested their method on \simtosim robot locomotion tasks.
It remains unclear whether this approach scales to \simtoreal scenarios since the adaptation is done based on the return obtained in simulation, thus is not physically grounded.
Bootstrapping from pre-recorded motion capture data of animals, \citet{Peng_Levine_20} learned quadruped locomotion skills with a synthesis of imitation learning, domain randomization, and domain adaptation (Section~\ref{sec_sota_transfer_learning}).
The introduced method is conceptually related to the approach of \citet{Yu_Turk_19}, but adds an information bottleneck.
%
%
According to the authors, this bottleneck is necessary because without it, the policy has access to the underlying dynamics parameters and becomes overly dependent on them, which leads to brittle behavior.
To avoid this overfitting, \citet{Peng_Levine_20} limit the mutual information between the domain parameters and their encoding, realized as penalty on the \ac{KL} divergence from a zero-mean Gaussian prior on the latent variable.


\paragraph*{The Bilevel Optimization Perspective}
\citet{Muratore_Peters_21_RAL} formulated adaptive domain randomization as a bilevel optimization that consists of an upper and a lower level problem.
In this framework, the upper level is concerned with finding the domain parameter distribution, which when used for training in simulation leads to a policy with maximal real-world return.
The lower level problem seeks to find a policy in the current randomized source domain.
Using \ac{BO} for the upper level and model-free \ac{RL} for the lower level, \citet{Muratore_Peters_21_RAL} compare their method in two underactuated \simtoreal robotic tasks against two baselines.
Picturing the real-world return analogous to the probability for optimality, this approach reveals parallels to control as inference~\citep{Rawlik_Vijayakumar_13,Levine_Koltun_13,Watson_Peters_21}, where the control variates are the parameters of the domain distribution.
\ac{BO} has also been employed by \citet{Paul_etal_2018} to adapt the distribution of domain parameters such that using these for the subsequent training maximizes the policy's return.
Their method models the relation between the current domain parameters, the current policy and the return of the updated policy with a \ac{GP}.
Choosing the domain parameters that maximize the return in simulation is critical, since this creates the possibility to adapt the environment such that it is easier for the agent to solve.
This design decision requires the policy parameters to be fed into the \ac{GP} which is prohibitively expensive if the full set of parameters are used.
Therefore, abstractions of the policy, so-called fingerprints, are created.
These handcrafted features, e.g., a Gaussian approximation of the stationary state distribution, replace the policy to reduce the input dimension.
\citet{Paul_etal_2018} tested the suggested algorithm on three \simtosim tasks, focusing on the handling of so-called significant rare events.
Embedding the domain parameters into the mean function of a \ac{GP} which models the system dynamics, \citet{Chatzilygeroudis_Mouret_18} extended a black-box policy search algorithm \citep{Chatzilygeroudis_Mouret_17} with a simulator as prior.
The approach explicitly searches for parameters of the simulator that fit the real-world data in an upper level loop, while optimizing the \ac{GP}'s hyper-parameters in a lower level loop.
This method allowed a damage hexapod robot to walk in less than 30 seconds.
\citet{Ruiz_Chandraker_19} proposed a meta-algorithm which is based on a bilevel optimization problem and updates the domain parameter distribution using REINFORCE~\citep{Williams_92}.
The approach has been evaluated in simulation on synthetic data, except for a semantic segmentation task.
Thus, there was no dynamics-dependent interaction of the learned policy with the real world. 
\citet{Mehta_Paull_19} also formulated the adaption of the domain parameter distribution as an \ac{RL} problem where different simulation instances are sampled and compared against a reference environment based on the resulting trajectories.
This comparison is done by a discriminator which yields rewards proportional to the difficulty of distinguishing the simulated and real environments, hence providing an incentive to generate distinct domains.
Using this reward signal, the domain parameters of the simulation instances are updated via \acl{SVPG}~\citep{Liu_etal_2017}.
\citet{Mehta_Paull_19} evaluated their method in a \simtoreal experiment where a robotic arm had to reach a desired point.
In contrast, \citet{Chebotar_Fox_19} presented a trajectory-based framework for closing the reality gap, and validated it on two \simtoreal robotic manipulation tasks.
The proposed procedure adapts the domain parameter distribution's parameters by minimizing discrepancy between observations from the real-world system and the simulation.
To measure the discrepancy, \citet{Chebotar_Fox_19} use a linear combination of the $L^1$ and $L^2$ norm between simulated and real trajectories.
These values are then plugged in as costs for \ac{REPS}~\citep{Peters_etal_2010} to update the simulator's parameters, hence turning the simulator identification into an episodic \ac{RL} problem.
The policy optimization was done using \ac{PPO}~\citep{Schulman_Klimov_17}, a step-based model-free \ac{RL} algorithm.

\paragraph*{Removing Restrictions on the Domain Parameter Distribution}
\citet{Ramos_Fox_19} perform a fully Bayesian treatment of the simulator's parameters by employing \acf{LFI} with a \ac{MDN} as model for the density estimator.
Analyzing the obtained posterior over domain parameters, they showed that the proposed method is, in a \simtosim scenario, able to simultaneously infer different parameter configurations which can explain the observed trajectories.
An evaluation over a gird of domain parameters confirms that the policies trained with the inferred posterior are more robust model uncertainties.
The key benefit over previous approaches is that the domain parameter distribution is not restricted to belong to a specific family, e.g., normal or uniform.
Instead, the true posterior is approximated by the density estimator, fitted using \ac{LFI}~\citep{Papamakarios_Murray_16}.
In follow-up work, \citet{Possas_Ramos_20} addressed the problem of learning the behavioral policies which are required for the collection of target domain data.
By describing the integration policy optimization via model-free \ac{RL}, the authors created an online variant of the original method.
The \simtoreal experiments were carried out using \ac{MPC} where (only) the model parameters are updated based on the result from the \ac{LFI} routine.
\citet{Matl_Fox_20} scaled the Bayesian inference procedure of \citet{Ramos_Fox_19} to the simulation of granular media, estimating parameters such as friction and restitution coefficients.
\citet{Barcelos_Ramos_20} presented a method that interleaves domain randomization, \ac{LFI}, and policy optimization.
The controller is updated via nonlinear \ac{MPC} while using the unscented transform to simulate different domain instances for the control horizon.
Hence, this algorithm allows one to calibrate the uncertainty as the system evolves with the passage of time, attributing higher costs to more uncertain paths.
For performing the essential \ac{LFI}, the authors build upon the work of \citet{Ramos_Fox_19} to identify the posterior domain parameters, which are modeled by a mixture of Gaussians. 
The approach was validated on a simulated inverted pendulum swing-up task as well as a real trajectory following task using a wheeled robot.
Since the density estimation problem is the center piece of \ac{LFI}-based domain randomization, improving the estimator's flexibility is of great interest.
\citet{Muratore_Peters_21_CoRL} employed a sequential neural posterior estimation algorithm~\citep{Greenberg_Macke_19} which uses normalizing flows to estimate the (conditional) posterior over simulators.
In combination with a segment-wise synchronization between the simulations and the recorded real-world trajectories, \citet{Muratore_Peters_21_CoRL} demonstrated the neural inference method's ability to learn the posterior belief over contact-rich black-box simulations.
Moreover, the proposed approach was evaluated with policy optimization in the loop on an underactuated swing-up and balancing task, showing improved results compared to BayesSim~\citep{Ramos_Fox_19} as well as Bayesian linear regression.

\subsection{Adversarial Domain Randomization}
\label{sec_sota_adversarial_dr}
	\begin{wrapfigure}[10]{r}{0.50\columnwidth} 
		\raggedleft
		\vspace*{-1.4\baselineskip}
		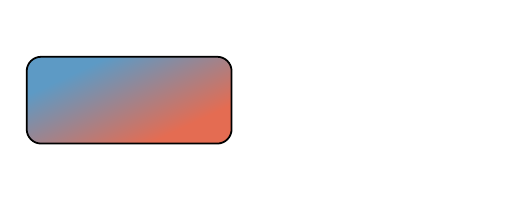
		\vspace*{-1.\baselineskip}
	\caption{Conceptual illustration of adversarial domain randomization.}
	\label{fig_sota_adversarial_dr}
	\end{wrapfigure}
Extensive prior studies have shown that deep \ac{NN} classifiers are vulnerable to imperceptible perturbations their inputs, obtained via adversarial optimization, leading to significant drops in accuracy~\citep{Szegedy_Fergus_13,Goodfellow_Szegedy_14,Fawzi_Frossard_15,Kurakin_Bengio_17,Ilyas_Madry_19}. 
This line of research has been extended to reinforcement learning, showing that small (adversarial) perturbations are enough to significantly degrade the policy performance~\citep{Huang_Abbeel_17}.
To defend against such attacks, the training data can be augmented with adversarially-perturbed examples, or the adversarial inputs can be detected and neutralized at test-time.
However, studies of existing defenses have shown that adversarial examples are harder to detect than originally believed~\citep{Carlini_Wagner_17}.
It is safe to assume that this insight gained from computer vision problems transfers to the \ac{RL} setting, on which we focus here.

\paragraph*{Adversary Available Analytically}
\citet{Mandlekar_Savarese_17} proposed physically plausible perturbations by randomly deciding when to add a scaled gradient of the expected return \wrt the state.
Their \simtosim evaluation on four MuJoCo tasks showed that agents trained with the suggested adversarial randomization generalize slightly better to domain parameter configurations than agents trained with a static randomization scheme.
\citet{Lutter_Garg_21_RSS} derived the optimal policy together with different optimal disturbances from the value function in a continuous state, action, and time \ac{RL} setting.
Despite outstanding \simtoreal transferability of the resulting policies, the presented approach is conceptually restricted by assuming access to a compact representation of the state domain, typically obtained through exhaustive sampling, which hinders the scalability to high-dimensional tasks.

\paragraph*{Adversary Learned via Two-Player Games}
Domain randomization can be described using a game theoretic framework.
Focusing on two-player games for model-based \ac{RL}, \citet{Rajeswaran_Kumar_20} define a \enquote*{policy player} which maximizes rewards in the learned model and a \enquote*{model player} which minimizes prediction error of data collected by policy player.
This formulation can be transferred to the \simtoreal scenario in different ways.
One example is to make the \enquote*{policy player} model-agnostic and to let the \enquote*{model player} control the domain parameters.
\citet{Pinto_Gupta_17} introduced the idea of a second agent whose goal it is to hinder the first agent from fulfilling its task.
This adversary has the ability to apply force disturbances at predefined locations of the robot's body, while the domain parameters remain unchanged.
Both agents are trained in alternation using \ac{RL} make this a zero-sum game.
Similarly, \citet{Zhang_Chen_Hsieh_21} aim to train an agent using adversarial examples such that it becomes robust against test-time attacks. 
As in the approach presented by \citet{Pinto_Gupta_17}, the adversary and the protagonist are trained alternately until convergence at every meta-iteration.
Unlike prior work, \citet{Zhang_Chen_Hsieh_21} build on state-adversarial \acp{MDP} manipulating the observations but not the simulation state.
Another key property of their approach is that the perturbations are applied after a projection to a bounded set.
The proposed observation-based attack as well as training algorithm is supported by four \simtosim validations in MuJoCo environments.
\citet{Jiang_Tan_21} employed \acp{GAN} to distinguish between source and target domain dynamics, sharing the concept of a learned domain discriminator with \citet{Mehta_Paull_19}.
Moreover, the authors proposed to augment an analytical physics simulator with a \ac{NN} that is trained to maximize the similarity between simulated and real trajectories, turning the identification of the hybrid simulator into an \ac{RL} problem.
The comparison on a \simtoreal quadruped locomotion task showed an advantage over static domain randomization baselines.
On the other hand, this method added noise to the behavioral policy in order to obtain diverse target domain trajectories for the simulator identification, which can be considered dangerous.

\section{Discussion and Outlook}
\label{sec_sota_discussion_and_outlook}
To conclude this review, we discuss practical aspects of choosing among the existing domain randomization approaches (Section~\ref{sec_sota_expectation_management}), emphasizing that \simtoreal transfer can also be achieved without randomizing (Section~\ref{sec_sota_transfer_wo_domain_randomization}).
Finally, we sketch out several promising directions for future \simtoreal research (Section~\ref{sec_sota_future_research}).

\subsection{Choosing a Suitable Domain Randomization Approach}
\label{sec_sota_expectation_management}
Every publication on \simtoreal robot learning presents an approach that surpasses its baselines.
So, how should we select the right algorithm given a task?
Up to now, there is no benchmark for \simtoreal methods based on the policy's target domain performance, and it is highly questionable if such a comparison could be fair, given that these algorithms have substantially different requirements and goals.
The absence of one common benchmark is not necessarily bad, since bundling a set of environments to define a metric would bias research to pursue methods which optimize solely for that metric.
A prominent example for this mechanism is the OpenAI Gym~\citep{Brockman_Zaremba_16}, which became the de facto standard for \ac{RL}. 
Contrarily, a similar development for \simtoreal research is not desirable since the overfitting to a small set of scenarios would be detrimental to the desired transferability and the vast amount of other scenarios.

When choosing from the published algorithms, the practitioner is advised to check if the approach has been tested on at least two different \simtoreal tasks, and if the (sometimes implicit) assumptions can be met.
Adaptive domain randomization methods, for example, will require operating the physical device in order to collect real-world data.
After all, we can expect that approaches with randomization will be more robust than the ones only trained on a nominal model.
This has been shown consistently (Sections~\ref{sec_sota_domain_randomization}).
However, we can not expect that these approaches work out of the box on novel problems without adjusting the hyper-parameters.
Another starting point could be the set of \simtosim benchmarks released by \citet{Mehta_Ramos_20}, targeting the problem of system identification for \sota domain randomization algorithms.

\subsection{\SimtoReal Transfer without Domain Randomization}
\label{sec_sota_transfer_wo_domain_randomization}
Domain randomization is one way to successfully transfer control policies learned in simulation to the physical device, but by no means the only way.

\paragraph*{Action Transformation}
In order to cope with the inaccuracies of a simulator, \citet{Christiano_Zaremba_16} propose to train a deep inverse dynamics model to map the action commanded by policy to a transformed action. When applying the original action to the real system and the transformed action to the simulated system, they would lead to the same next robot state, thus bridging the reality gap.
To generate the data for training the inverse dynamics model, preliminary policies are augmented with hand-tuned exploration noise and executed in the target domain.
Their approach is based on the observation that a policy's high-level strategy remains valid after \simtoreal transfer, and assumes that the simulator provides a reasonable estimate of the next state.
With the same goal in mind, \citet{Hanna_Stone_17} suggest an action transformation that is learned such that applying the transformed actions in simulation has the same effects as applying the original actions had on the real system.
At the core approach is the estimation of neural forward and inverse models based on rollouts executed with the real robot.
%

\paragraph*{Novel Neural Policy Architectures}
\citet{Rusu_Hadsell_17} employ a progressively growing \ac{NN} architecture~\citep{Rusu_Hadsell_16} to learn an end-to-end approach mapping from pixels to discretized joint velocities. 
This \ac{NN} framework enables the reuse of previously gained knowledge as well as the adaptation to new input modalities.
The first part of the \ac{NN} policy is trained in simulation, while the part added when transferring needs to be trained using real-world data.
For a relatively simple reaching task, the authors reported requiring approximately four hours of runtime on the physical robot.

\paragraph*{Identifying and Improving the Simulator}
\citet{Xie_VanDePanne_19} describe an iterative process including motion tracking, system identification, \ac{RL}, and knowledge distillation, to learn control policies for humanoid walking on the physical system.
This way, the authors can rely on known building blocks resulting in initial and intermediate policies which are reasonably safe to execute.
To run a policy on the real robot while learning without the risk of damaging or stopping the device, \citet{Kaspar_Bock_20} propose to combine operational space control and \ac{RL}.
After carefully identifying the simulator's parameters, the \ac{RL} agent learns to control the end-effector via forces on a unit mass-spring-damper system.
The constrains and nullspace behavior are abstracted away from the agent, making the \ac{RL} problem easier and the policy more transferable.
%


\subsection{Promising Future Research Directions}
\label{sec_sota_future_research}
Learning from randomized simulations still offers abundant possibilities to enable or improve the \simtoreal transfer of control policies.
In the following section, we describe multiple opportunities for future work in this area of research.

\paragraph*{Real-to-Sim-to-Real Transfer}
Creating randomizable simulation environments is time-intensive, and the initial guesses for the domain parameters as well as their variances are typically very inaccurate.
It is of great interest to automate this process grounded by real-world data.
One viable scenario could be to record an environment with a \acs{RGBD} camera, and subsequently use the information to reconstruct the scene.
Moreover, the recorded data can be processed to infer the domain parameters, which then specifies the domain parameter distributions.
When devising such a framework, we could start from prior work on 3D scene reconstruction \cite{Kolev_etal_2009,Haefner_etal_2018} as well as methods to estimate the degrees of freedom for rigid bodies~\citep{MartinMartin_Brock_2014}.
A data-based automatic generation of simulation environments (real-to-\simtoreal) not only promises to reduce the workload, but would also yields a meaningful initialization for domain distribution parameters.

\paragraph*{Policy Architectures with Inductive Biases}
Deep \acp{NN} are by far the most common policy type, favored because of their flexibility and expressiveness. 
However, they are also brittle \wrt changes in their inputs~\citep{Szegedy_Fergus_13,Goodfellow_Szegedy_14,Huang_Abbeel_17}.
Due to the inevitable domain shift in \simtoreal scenarios this input sensitivity is magnified.
The success of domain randomization methods for robot learning can largely be attributed to their ability of regularizing deep \ac{NN} policies by diversifying the training data.
Generally, one may also introduce regularization to the learning by designing alternative models for the control policies, e.g., linear combination of features and parameters, (time varying) mixtures of densities, or movement primitives.
All of these have their individual strengths and weaknesses.
We believe that pairing the expressiveness of deep \acp{NN} with physically-grounded prior knowledge leads to controllers that achieve high performance and suffer less from transferring to the real world, since they are able to bootstrap from their prior.
There are multiple ways to incorporate abstract knowledge about physics. 
We can for example restrict the policy to obey stable system dynamics derived from first principles~\citep{Greydanus_Yosinski_19,Lutter_Peters_19}.
Another approach is to design the model class such that the closed-loop system is passive for all parameterizations of the learned policy, thus guaranteeing stability in the sense of Lyapunov as well as bounded output energy given bounded input energy~\citep{Brogliato_07,Yang_Ma_13,Dai_Tedrake_21}. All these methods would require significant exploration in the environment, making it even more challenging to learn successful controllers in the real-world directly. Leveraging randomized simulation is likely going to be a critical component in demonstrating solving sequential problems on real robots.

\paragraph*{Towards Dual Control via Neural \acl{LFI}}
Continuing the direction of adaptive domain randomization, we are convinced that neural \ac{LFI} powered by normalizing flows are auspicious approaches.
The combination of highly flexible density estimators with widely applicable and sample-efficient inference methods allows one to identify multi-modal distributions over simulators with very mild assumptions~\citep{Ramos_Fox_19,Barcelos_Ramos_21,Muratore_Peters_21_CoRL}.
By introducing an auxiliary optimality variable and making the policy parameters subject to the inference, we obtain the posterior over policies quantifying their likelihood of being optimal.
While this idea is well-known in the control-as-inference community~\citep{Rawlik_Vijayakumar_13,Levine_Koltun_13,Watson_Peters_21}, prior methods were limited to less powerful density estimation procedures.
Taking this idea one step further, we could additionally include the domain parameters for inference, and thereby establish connections to dual control~\citep{Feldbaum_60,Wittenmark_95}.

\paragraph*{Accounting for the Cost of Information Collection}
Another promising direction for future research is the combination of simulated and real-world data collection with explicit consideration of the different costs when sampling from the two domains, subject to a restriction of the overall computational budget.
One part of this problem was already addressed by \citet{Marco_Trimpe_17}, showing how simulation can be used to alleviate the need for real-world samples when finding a set of policy parameters.
However, the question of how to schedule the individual (simulated or real) experiments and when to stop the procedure, i.e., when does the cost of gathering information exceed its expected benefit, is not answered for \simtoreal transfer yet.
This question relates to the problems of optimal stopping~\citep{Chow_Robbins_63} as well as multi-fidelity optimization~\citep{Forrester_Keane_07}, and can be seen as a reformulation thereof in the context of simulation-based learning.

\paragraph*{Solving Sequential Problems}
The problem settings considered in the overwhelming majority of related publications, are (continuous) control tasks which do not have a sequential nature.
In contrast, most real-world tasks such as the ones posed at the DARPA Robotics Challenge~\citep{Krotkov_Orlowski_17} consist of (disconnected) segments, e.g., a robot needs to turn the knob before it can open a door.
One possible way to address these more complicated tasks is by splitting the control into high and low level policies, similar to the options framework~\citep{Sutton_Singh_99}.
The higher level policy is trained to orchestrate the low-level policies which could be learned or fixed.
Existing approaches typically realize this with discrete switches between the low-level policies, leading to undesirable abrupt changes in the behavior.
An alternative would be a continuous blending of policies, controlled by a special kind of recurrent \ac{NN} which has originally been proposed by \citet{Amari_77} to model activities in the human brain.
Used as policy architectures they can be constructed to exhibit asymptotically stable nonlinear dynamics~\citep{Kishimoto_Amari_79}.
The main benefits of this structure are its easy interpretability via exhibition and inhibition of neural potentials, as well as the relatively low number of parameters necessary to create complex and adaptive behavior.
A variation of this idea with hand-tuned parameters, i.e., without machine learning, has been applied by \citet{Luksch_Yoshiike_12} to coordinate the activation pre-defined movement primitives.

\section*{Selection of References}
We chose the references based on multiple criteria:
(i) Our primary goal was to covering all milestones of the \simtoreal research for robotics.
(ii) In the process, we aimed at diversifying over subfields and research groups.
(iii) A large proportion of papers came to our attention by running Google Scholar alerts on \enquote{\simtoreal} and \enquote{reality gap} since 2017.
(iv) Another source were reverse searches starting from highly influential publications.
(v) Some papers came to our attention because of citation notifications we received on our work.
(vi) Finally, a few of the selected publications are recommendations from reviewers, colleagues, or researchers met at conferences. 
(vii) Peer-reviewed papers were strongly preferred over pre-prints.

\section*{Funding}
Fabio~Muratore gratefully acknowledges the financial support from \acl{HRIE}.
Jan~Peters received funding from the European Union’s Horizon 2020 research and innovation programme under grant agreement No 640554. Wenhao~Yu and Greg~Turk have been supported by NSF award IIS-1514258.

\section*{Conflict of Interest Statement}
Author Fabio Muratore is employed by the Technical University of Darmstadt in collaboration with the \acl{HRIE}. The remaining authors declare that the research was conducted in the absence of any commercial or financial relationships that could be construed as a potential conflict of interest.

\bibliographystyle{frontiersinSCNS_ENG_HUMS} 
\IfFileExists{\string~/HESSENBOX-DA/fMRT_LaTeX/Bibliography/fMRT.bib}%
{\bibliography{\string~/HESSENBOX-DA/fMRT_LaTeX/Bibliography/fMRT.bib}}
{\bibliography{fMRT.bib}}

\end{document}

%% file: fMRT_acronyms.tex
\acrodef{AALR}{Amortized Approximate Likelihood Ratio}
\acrodef{ABC}{Approximate Bayesian Computation}
\acrodef{ADaPT}{Adaptive Policy Transfer for Stochastic Dynamics}
\acrodef{ADN}{Activation Dynamics Network}
\acrodef{ADR}{Active Domain Randomization}
\acrodef{ARC-t}{Asymmetric Regularized Cross-domain transformation}
\acrodef{A2RP}{Averaged Two-Replication Procedure}
\acrodef{APT}{Automatic Posterior Transformation}
\acrodef{ARPL}{Adversarially Robust Policy Learning}
\acrodef{BO}{Bayesian Optimization}
\acrodef{BayRn}{Bayesian Domain Randomization}
\acrodef{CAD}{Computer Aided Design}
\acrodef{CIO}{Contact-Invariant Optimization}
\acrodef{CMA}{Covariance Matrix Adaptation}
\acrodef{CMA-ES}{Covariance Matrix Adaptation - Evolutionary Strategies}
\acrodef{CMDP}{Contextual Markov Decision Process}
\acrodefplural{CMPD}[CMPDs]{Contextual Markov Decision Processes}
\acrodef{CNN}{Convolutional Neural Network}
\acrodef{CoM}{Center of Mass}
\acrodef{CVaR}{Conditional Value at Risk}
\acrodef{DA}{Domain Adaptation}
\acrodef{DAN}{Deep Adaptation Network}
\acrodef{DDPG}{Deep Deterministic Policy Gradient}
\acrodef{DISCO}{Double likelihood-free Inference Stochastic COntrol}
\acrodef{DMP}{Dynamic Movement Primitive}
\acrodefplural{DMP}[DMPs]{Dynamic Movement Primitives}
\acrodef{DNN}{Deep Neural Network}
\acrodef{DoF}{Degree of Freedom}
\acrodefplural{DOF}[DOF]{Degrees of Freedom}
\acrodef{DR}{Domain Randomization}
\acrodef{DRL}{Deep Reinforcement Learning}
\acrodef{DTW}{Dynamic Time Warping}
\acrodef{DS}{Dynamical System}
\acrodef{EoM}{Equation of Motion}
\acrodefplural{EoM}[EoMs]{Equations of Motion}
\acrodef{EEA}{Estimation Exploration Algorithm}
\acrodef{EPOpt}{Ensemble Policy Optimization}
\acrodef{FNN}{Feedforward Neural Network}
\acrodef{FPO}{Fingerprint Policy Optimization}
\acrodef{GA}{Genetic Algorithm}
\acrodef{GAN}{Generative Adversarial Network}
\acrodef{GMM}{Gaussian Mixture Model}
\acrodef{GP}{Gaussian Process}
\acrodefplural{GP}[GPs]{Gaussian Processes}
\acrodef{GPS}{Guided Policy Search}
\acrodef{GPU}{Graphics Processing Unit}
\acrodef{HER}{Hindsight Experience Replay}
\acrodef{HRI}{Honda Research Institute}
\acrodef{HRIE}{Honda Research Institute Europe}
\acrodef{I2RP}{Independent Two-Replication Procedure}
\acrodef{IAS}{Intelligent Autonomous Systems}
\acrodef{IEEE}{Institute of Electrical and Electronics Engineers}
\acrodef{IID}{Independent and Identically Distributed}
\acrodef{KL}{Kullback-Leibler}
\acrodef{LFI}{Likelihood-Free Inference}
\acrodef{LTI}{Linear Time-Invariant}
\acrodef{LQR}{Linear-Quadratic Regulator}
\acrodef{LSDA}{Large Scale Detection through Adaptation}
\acrodef{LSTM}{Long Short-Term Memory}
\acrodef{LWPR}{Locally Weighted Projection Regression}
\acrodef{MAF}{Masked Autoregressive Flow}
\acrodef{MCMC}{Markov Chain Monte Carlo}
\acrodef{MDN}{Mixture Density Network}
\acrodef{MDP}{Markov Decision Process}
\acrodefplural{MDP}{Markov Decision Processes}
\acrodef{MIAC}{Model Identification Adaptive Control}
\acrodef{MoG}{Mixture of Gaussians}
\acrodef{MP}{Movement Primitive}
\acrodef{MPC}{Model Predictive Control}
\acrodef{MPPI}{Model Predictive Path Integral}
\acrodef{MSE}{Mean-Squared Error}
\acrodef{NN}{Neural Network}
\acrodef{NPDR}{Neural Posterior Domain Randomization}
\acrodef{NPG}{Natural Policy Gradient}
\acrodef{ODE}{Ordinary Differential Equation}
\acrodef{ODEphys}[ODE]{Open Dynamics Engine}
\acrodef{OG}{Optimality Gap}
\acrodef{PNN}{Progressive Neural Network}
\acrodef{PD}{Proportional-Derivative}
\acrodef{POMDP}{Partially-Observable Markov Decision Process}
\acrodef{PoWER}{Policy learning by Weighting Exploration with the Returns}
\acrodef{PPO}{Proximal Policy Optimization}
\acrodef{ProMP}{Probabilistic Movement Primitive}
\acrodef{QBB}{Quanser Ball-Balancer}
\acrodef{QCP}{Quanser Cart-Pole}
\acrodef{QQ}{Quanser Qube}
\acrodef{RA}{Retrospective Approximation}
\acrodef{RARL}{Robust Adversarial Reinforcement Learning}
\acrodef{RBF}{Radial Basis Function}
\acrodef{REPS}{Relative Entropy Policy Search}
\acrodef{ResNet}{Residual neural network}
\acrodef{RFF}{Random Fourier Features}
\acrodef{RGB}{Red-Green-Blue}
\acrodef{RGBD}{Red-Green-Blue-Depth}
\acrodef{RMSE}{Root-Mean-Square Error}
\acrodef{RNN}{Recurrent Neural Network}
\acrodef{RL}{Reinforcement Learning}
\acrodef{SAA}{Sample Average Approximation}
\acrodef{SMC}{Sequential Monte Carlo}
\acrodef{SNLE}{Sequential Neural Likelihood Estimation}
\acrodef{SNPE}{Sequential Neural Posterior Estimation}
\acrodef{SNRE}{Sequential Neural Ratio Estimation}
\acrodef{SOB}{Simulation Optimization Bias}
\acrodef{SP}{Stochastic Program}
\acrodefplural{SP}{Stochastic Programs}
\acrodef{SPOTA}{Simulation-based Policy Optimization with Transferability Assessment}
\acrodef{SPRL}{Self-Paced Reinforcement Learning}
\acrodef{SRCC}{Sim-vs-Real Correlation Coefficient}
\acrodef{SVPG}{Stein Variational Policy Gradient}
\acrodef{SysId}{System Identification}
\acrodef{TRPO}{Trust Region Policy Optimization}
\acrodef{UCB}{Upper Confidence Bound}
\acrodef{UDR}{Uniform Domain Randomization}
\acrodef{UCBOG}{Upper Confidence Bound on the Optimality Gap}
\acrodef{UCSOB}{Upper Confidence bound on the Simulation Optimization Bias}
\acrodef{UP-OSI}{Universal Policy - Online System Identification}
\acrodef{VAE}{Variational AutoEncoder}

%% file: fMRT_macros.tex

\makeatletter
\@ifpackageloaded{xparse}{}{\usepackage{xparse}}
\@ifpackageloaded{tabularx}{}{\usepackage{tabularx}} 
\@ifpackageloaded{xifthen}{}{\usepackage{xifthen}}
\@ifpackageloaded{xspace}{}{\usepackage{xspace}} 
\@ifpackageloaded{amsopn}{}{\usepackage{amsopn}} 
\@ifpackageloaded{amssymb}{}{\usepackage{amssymb}} 
\@ifpackageloaded{mathtools}{}{\usepackage{mathtools}} 
\@ifpackageloaded{bm}{}{\usepackage{bm}} 
\@ifpackageloaded{siunitx}{}{%
\usepackage[detect-all,scientific-notation=false,separate-uncertainty=true]{siunitx} 
\sisetup{%
    retain-explicit-plus=true,
    detect-weight=true,
    detect-family=true,
	output-exponent-marker=\ensuremath{\mathrm{e}},
	per-mode=symbol,
	range-phrase={,}\ ,
    range-units=brackets,
    open-bracket={[},
    close-bracket=],
}%
}
\makeatother

\definecolor{fMRTblack}{HTML}{2B2E34}
\definecolor{fMRTwhite}{HTML}{F3EEE1}
\definecolor{fMRTyellow}{HTML}{FFD564}
\definecolor{fMRTorange}{HTML}{FF5500}
\colorlet{fMRTlightGray}{white!80!fMRTblack}
\colorlet{fMRTgray}{white!35!fMRTblack}
\colorlet{fMRTdarkGray}{white!20!fMRTblack}
\colorlet{fMRTlightOrange}{fMRTorange!60!fMRTwhite}
\definecolor{fMRTblue}{HTML}{3333B3}
\colorlet{fMRTlightBlue}{fMRTblue!30!white}
\definecolor{fMRTverylightRed}{HTML}{FF8E6B}
\colorlet{fMRTlightRed}{red!60!fMRTwhite}
\colorlet{fMRTdarkRed}{red!70!fMRTgray} 
\definecolor{fMRTlightGreen}{HTML}{8CDD81} 
\definecolor{fMRTborwn}{HTML}{8B4513} 
\colorlet{fMRTlightBorwn}{fMRTborwn!60!fMRTwhite}

\colorlet{fMRTdark@backgroundInner}{fMRTblack}
\colorlet{fMRTdark@backgroundOuter}{fMRTblack}
\colorlet{fMRTlight@backgroundInner}{fMRTwhite}
\colorlet{fMRTlight@backgroundOuter}{fMRTwhite}




\newcolumntype{L}{>{\raggedright\arraybackslash}X}
\newcolumntype{R}{>{\raggedleft\arraybackslash}X}
\newcolumntype{C}{>{\centering\arraybackslash}X}
\newcolumntype{P}[1]{>{\raggedright\arraybackslash}p{#1}} 



\newcommand{\optsym}{\star} 


\newcommand{\realsym}{\text{real}}
\newcommand{\real}{^\realsym}

\newcommand{\RR}{\mathbb{R}} 

\DeclareDocumentCommand{\set}{O{} O{} m}{\{#3\}_{#1}^{#2}} 
\DeclareDocumentCommand{\tuple}{O{} O{} m}{\left\langle#3\right\rangle_{#1}^{#2}} 

%
\newcommand{\stateset}[1][]{\mathcal{S}_{#1}} 
\newcommand{\obsset}[1][]{\mathcal{O}_{#1}} 
\newcommand{\actionset}[1][]{\mathcal{A}_{#1}} 
\newcommand{\transprobsym}{\mathcal{P}}
\makeatletter
\DeclareDocumentCommand{\@transprob}{O{} O{} m}{\transprobsym_{#1}^{#2}\left( #3\right) }  
\DeclareDocumentCommand{\@@transprob}{O{} O{} m}{\transprobsym_{#1}^{#2}\!\left( #3\right) }
\newcommand{\transprob}{\@ifstar\@transprob\@@transprob}
\makeatother
\newcommand{\initstatedistrsym}{\mu}
\makeatletter
\DeclareDocumentCommand{\@initstatedistr}{O{} O{} m}{\initstatedistrsym_{#1}^{#2}\left( #3\right) }  
\DeclareDocumentCommand{\@@initstatedistr}{O{} O{} m}{\initstatedistrsym_{#1}^{#2}\!\left( #3\right) }
\newcommand{\initstatedistr}{\@ifstar\@initstatedistr\@@initstatedistr}
\makeatother
\newcommand{\mdp}[1][]{\mathcal{M}_{#1}} 
\newcommand{\dimstate}{{n_s}} 
\newcommand{\dimobs}{{n_o}} 
\newcommand{\dimact}{{n_a}} 
\newcommand{\dimpolparam}{{n_{\polparam}}} 
\newcommand{\dimdomparam}{{n_{\domparam}}} 




\newcommand{\wrt}{w.r.t.\xspace}
\newcommand{\aka}{a.k.a.\xspace}

\newcommand{\simtosim}{{sim-to-sim}\xspace}
\newcommand{\SimtoReal}{{Sim-to-Real}\xspace}

\newcommand{\simtoreal}{{sim-to-real}\xspace}
\newcommand{\sota}{{state-of-the-art}\xspace}

\newcommand{\bic}{ball-in-a-cup\xspace}


\newcommand{\Esub}[2]{\mathbb{E}_{#1} \! \left[ #2\right] } 

\makeatletter
\newcommand{\@KL}[2]{\mathrm{KL}(#1 \Vert #2)} 
\newcommand{\@@KL}[3][]{\mathrm{KL}\!\left( #2 #1\Vert #3\right)} 
\newcommand{\KL}{\@ifstar\@KL\@@KL}
\makeatother


\renewcommand{\exp}[1]{\mathrm{exp}\left( #1\right) } 


\makeatletter
\newcommand{\@given}[2]{\left. #1\right| #2} 
\newcommand{\@@given}[3][]{#2 #1| #3} 
\newcommand{\given}{\@ifstar\@given\@@given}
\makeatother


\DeclareDocumentCommand{\distmeas}{O{} O{} m}{\text{D}_{#1}^{#2} \!\left( #3\right)} 





\newcommand{\rewsym}{r}
\DeclareDocumentCommand{\rewfcn}{O{} O{} m}{\rewsym_{#1}^{#2}\!\left( #3\right) }


\newcommand{\trajssym}{\bm{\tau}}
\newcommand{\trajspace}{\mathcal{T}}

\DeclareDocumentCommand{\traj}{O{} O{}}{\trajssym_{#1}^{#2}} 
\newcommand{\statedistrsym}{\mu}
\DeclareDocumentCommand{\statedistr}{O{} O{} m}{\statedistrsym_{#1}^{#2}\!\left( #3\right) } 

\newcommand{\domparam}{\xi}
\newcommand{\domparams}{\bm{\xi}}

\newcommand{\domparamdistrsym}{p} 
\makeatletter
\DeclareDocumentCommand{\@domparamdistr}{O{} O{} m}{\domparamdistrsym_{#1}^{#2}\left( #3\right) }  
\DeclareDocumentCommand{\@@domparamdistr}{O{} O{} m}{\domparamdistrsym_{#1}^{#2}\!\left( #3\right) }
\newcommand{\domparamdistr}{\@ifstar\@domparamdistr\@@domparamdistr}
\makeatother


\newcommand{\domdistrparams}{\bm{\phi}} 

\newcommand{\densestsym}{q}
\DeclareDocumentCommand{\densest}{O{} O{} m}{\densestsym_{#1}^{#2}\!\left( #3\right)} 


\DeclareDocumentCommand{\domparamprior}{O{} O{} m}{\domparamdistrsym_{#1}^{#2}\!\left( #3\right) } 
\DeclareDocumentCommand{\domparamproposal}{O{} O{} m}{\tilde{\domparamdistrsym}_{#1}^{#2}\!\left( #3\right) } 
\DeclareDocumentCommand{\domparamlikelihood}{O{} O{} m}{\domparamdistrsym_{#1}^{#2}\!\left( #3\right) } 
\DeclareDocumentCommand{\domparamposterior}{O{} O{} m}{\domparamdistrsym_{#1}^{#2}\!\left( #3\right) } 
\DeclareDocumentCommand{\estdomparamposterior}{O{} O{} m}{\hat{\domparamdistrsym}_{#1}^{#2}\!\left( #3\right) } 

\newcommand{\genmodelsym}{g}
\DeclareDocumentCommand{\genmodel}{O{} O{} m}{\genmodelsym{#1}^{#2}\!\left( #3\right) }

\newcommand{\summstatsym}{f}

\DeclareDocumentCommand{\summstatfcn}{O{} O{} m}{\summstatsym_{#1}^{#2}\!\left( #3\right) }

\newcommand{\polsym}{\pi}
\DeclareDocumentCommand{\pol}{O{} O{} m}{\polsym_{#1}^{#2}\!\left( #3\right) } 
\newcommand{\polparam}{\theta}

\DeclareDocumentCommand{\polparams}{O{} O{}}{\ftheta_{#1}^{#2}} 
\DeclareDocumentCommand{\polparamsopt}{O{} O{}}{\ftheta_{#1}^{#2 \optsym}} 
\DeclareDocumentCommand{\polparamopt}{O{} O{}}{\polparam_{#1}^{#2 \optsym}} 


\newcommand{\acqfcnsym}{a}
\DeclareDocumentCommand{\acqfcn}{O{} O{} m}{\acqfcnsym_{#1}^{#2}\!\left( #3\right) }

\newcommand{\optgapsym}{G}
\DeclareDocumentCommand{\optgap}{O{} O{} m}{\optgapsym_{#1}^{#2}\!\left( #3\right) } 
\newcommand{\estoptgapsym}{\hat{G}}
\DeclareDocumentCommand{\estoptgap}{O{} O{} m}{\estoptgapsym_{#1}^{#2}\!\left( #3\right) } 
\newcommand{\meanoptgapsym}{\bar{G}}
\DeclareDocumentCommand{\optgapmean}{O{} O{} m}{\meanoptgapsym_{#1}^{#2}\!\left( #3\right) } 
\newcommand{\optgapsampsetsym}{\mathcal{G}}
\DeclareDocumentCommand{\optgapsampset}{O{} O{}}{\optgapsampsetsym_{#1}^{#2}} 

\iftrue
\newcommand{\bootsym}{B}
\DeclareDocumentCommand{\bootestoptgap}{O{} O{} m}{\estoptgapsym_{#1}^{\bootsym #2}\!\left( #3\right) }
\DeclareDocumentCommand{\bootmeanoptgap}{O{} O{} m}{\meanoptgapsym_{#1}^{\bootsym #2}\!\left( #3\right) }
\DeclareDocumentCommand{\bootoptgapsampset}{O{} O{}}{\optgapsampsetsym_{#1}^{\bootsym #2}}
\fi

\newcommand{\edrsym}{J}
\DeclareDocumentCommand{\edr}{O{} O{} m}{\edrsym_{#1}^{#2}\!\left( #3\right)} 

\newcommand{\estedrsym}{\hat{J}}
\DeclareDocumentCommand{\estedr}{O{}  O{} m}{\estedrsym_{#1}^{#2}\!\left( #3\right)} 
\newcommand{\eedrsym}{J} 
\DeclareDocumentCommand{\eedr}{O{} O{} m}{\eedrsym_{#1}^{#2}\!\left( #3\right)} 
\DeclareDocumentCommand{\esteedr}{O{} O{} m}{\estedrsym_{#1}^{#2}\!\left( #3\right)} 
\newcommand{\objfunsym}{f}
\DeclareDocumentCommand{\objfun}{O{} m}{\objfunsym_{#1}\!\left( #2\right)} 


\newcommand{\dsdynsym}{f}
\DeclareDocumentCommand{\dsdyn}{O{} m}{\dsdynsym_{#1}\!\left( #2\right)} 

\newcommand{\potdynsym}{f}
\newcommand{\potsdynsym}{\ff}
\DeclareDocumentCommand{\potdyn}{O{} m}{\potdynsym_{#1}\!\left( #2\right)} 
\DeclareDocumentCommand{\potsdyn}{O{} m}{\potsdynsym_{#1}\!\left( #2\right)} 


\ExplSyntaxOn
\NewDocumentCommand{\eqsref}{m}{\quinn_eqsref:n {#1}}
\seq_new:N \l_quinn_eqsref_seq
\cs_new:Npn \quinn_eqsref:n #1
{
	\seq_set_split:Nnn \l_quinn_eqsref_seq { , } { #1 }
	\seq_pop_right:NN \l_quinn_eqsref_seq \l_tmpa_tl
	( 
	\seq_map_inline:Nn \l_quinn_eqsref_seq
	{ \ref{##1},\nobreakspace } 
	\exp_args:NV \ref \l_tmpa_tl 
	) 
}
\ExplSyntaxOff

\SetKw{kwNot}{not}
\SetKw{kwAnd}{and}
\SetKw{kwBreak}{break}
\SetKwInOut{Input}{input}
\SetKwInOut{Output}{output}
\SetKwInOut{Initialization}{init}
\SetKwComment{Comment}{$\triangleright$\ }{}

\SetCommentSty{famuracommfont}
\SetKwRepeat{Do}{do}{while} 
\SetKwBlock{With}{with}{end} 
\newcommand{\llIf}[2]{{\let\par\relax\lIf{#1}{#2}}}
\newcommand{\llElse}[1]{{\let\par\relax\lElse{#1}}}
\newcommand{\llFor}[2]{{\let\par\relax\lFor{#1}{#2}}}


\DeclareBoldMathCommand{\fnull}{0}
\DeclareBoldMathCommand{\fO}{0}
\DeclareBoldMathCommand{\fA}{A}
\DeclareBoldMathCommand{\fa}{a}
\DeclareBoldMathCommand{\fB}{B}
\DeclareBoldMathCommand{\fb}{b}
\DeclareBoldMathCommand{\fC}{C}
\DeclareBoldMathCommand{\fc}{c}
\DeclareBoldMathCommand{\fD}{D}
\DeclareBoldMathCommand{\fd}{d}
\DeclareBoldMathCommand{\fE}{E}
\DeclareBoldMathCommand{\fe}{e}
\DeclareBoldMathCommand{\fF}{F}
\DeclareBoldMathCommand{\ff}{f}
\DeclareBoldMathCommand{\fG}{G}
\DeclareBoldMathCommand{\fg}{g}
\DeclareBoldMathCommand{\fH}{H}
\DeclareBoldMathCommand{\fh}{h}
\DeclareBoldMathCommand{\fI}{I}
\DeclareBoldMathCommand{\fJ}{J}
\DeclareBoldMathCommand{\fK}{K}
\DeclareBoldMathCommand{\fM}{M}
\DeclareBoldMathCommand{\fm}{m}
\DeclareBoldMathCommand{\fN}{N}
\DeclareBoldMathCommand{\fn}{n}
\DeclareBoldMathCommand{\fo}{o}
\DeclareBoldMathCommand{\fP}{P}
\DeclareBoldMathCommand{\fp}{p}
\DeclareBoldMathCommand{\fQ}{Q}
\DeclareBoldMathCommand{\fq}{q}
\DeclareBoldMathCommand{\fq}{q}

\DeclareBoldMathCommand{\fR}{R}
\DeclareBoldMathCommand{\fr}{r}

\DeclareBoldMathCommand{\fS}{S}
\DeclareBoldMathCommand{\fs}{s}

\DeclareBoldMathCommand{\ft}{t}
\DeclareBoldMathCommand{\fT}{T}
\DeclareBoldMathCommand{\fU}{U}
\DeclareBoldMathCommand{\fu}{u}
\DeclareBoldMathCommand{\fV}{V}
\DeclareBoldMathCommand{\fv}{v}
\DeclareBoldMathCommand{\fW}{W}
\DeclareBoldMathCommand{\fw}{w}
\DeclareBoldMathCommand{\fx}{x}
\DeclareBoldMathCommand{\fz}{z}

\DeclareBoldMathCommand{\fY}{Y}
\DeclareBoldMathCommand{\fy}{y}
\DeclareBoldMathCommand{\fZ}{Z}
\DeclareBoldMathCommand{\falpha}{\alpha}

\DeclareBoldMathCommand{\fbeta}{\beta}
\DeclareBoldMathCommand{\fchi}{\chi}
\DeclareBoldMathCommand{\fepsilon}{\epsilon}
\DeclareBoldMathCommand{\fvarepsilon}{\varepsilon}
\DeclareBoldMathCommand{\fgamma}{\gamma}
\DeclareBoldMathCommand{\fGamma}{\Gamma}
\DeclareBoldMathCommand{\flambda}{\lambda}
\DeclareBoldMathCommand{\fLambda}{\Lambda}
\DeclareBoldMathCommand{\fmu}{\mu}
\DeclareBoldMathCommand{\fnu}{\nu}
\DeclareBoldMathCommand{\fomega}{\omega}
\DeclareBoldMathCommand{\fpi}{\pi}
\DeclareBoldMathCommand{\fphi}{\phi}
\DeclareBoldMathCommand{\fPhi}{\Phi}
\DeclareBoldMathCommand{\fpsi}{\psi}
\DeclareBoldMathCommand{\fsigma}{\sigma}
\DeclareBoldMathCommand{\fSigma}{\Sigma}
\DeclareBoldMathCommand{\ftau}{\tau}
\DeclareBoldMathCommand{\ftheta}{\theta}
\DeclareBoldMathCommand{\fTheta}{\Theta}

\DeclareBoldMathCommand{\fxi}{\xi}

%% file: images/topology_sim2real_v2.pdf_tex
\begingroup%
  \makeatletter%
  \providecommand\color[2][]{%
    \errmessage{(Inkscape) Color is used for the text in Inkscape, but the package 'color.sty' is not loaded}%
    \renewcommand\color[2][]{}%
  }%
  \providecommand\transparent[1]{%
    \errmessage{(Inkscape) Transparency is used (non-zero) for the text in Inkscape, but the package 'transparent.sty' is not loaded}%
    \renewcommand\transparent[1]{}%
  }%
  \providecommand\rotatebox[2]{#2}%
  \newcommand*\fsize{\dimexpr\f@size pt\relax}%
  \newcommand*\lineheight[1]{\fontsize{\fsize}{#1\fsize}\selectfont}%
  \ifx\svgwidth\undefined%
    \setlength{\unitlength}{458.38052342bp}%
    \ifx\svgscale\undefined%
      \relax%
    \else%
      \setlength{\unitlength}{\unitlength * \real{\svgscale}}%
    \fi%
  \else%
    \setlength{\unitlength}{\svgwidth}%
  \fi%
  \global\let\svgwidth\undefined%
  \global\let\svgscale\undefined%
  \makeatother%
  \begin{picture}(1,0.19332242)%
    \lineheight{1}%
    \setlength\tabcolsep{0pt}%
    \put(0,0){\includegraphics[width=\unitlength,page=1]{topology_sim2real_v2.pdf}}%
    \put(0.49998057,0.09011652){\color[rgb]{0,0,0}\makebox(0,0)[t]{\lineheight{1.25}\smash{\begin{tabular}[t]{c}Sim-to-Real\end{tabular}}}}%
    \put(0,0){\includegraphics[width=\unitlength,page=2]{topology_sim2real_v2.pdf}}%
    \put(0.49998025,0.16045816){\color[rgb]{0,0,0}\makebox(0,0)[t]{\lineheight{1.25}\smash{\begin{tabular}[t]{c}\hyperref[sec_sota_meta_learning]{Meta Learning}\end{tabular}}}}%
    \put(0,0){\includegraphics[width=\unitlength,page=3]{topology_sim2real_v2.pdf}}%
    \put(0.16074574,0.09011646){\color[rgb]{0,0,0}\makebox(0,0)[t]{\lineheight{1.25}\smash{\begin{tabular}[t]{c}\hyperref[sec_sota_distributional_robustness]{Distributional Robustness}\end{tabular}}}}%
    \put(0,0){\includegraphics[width=\unitlength,page=4]{topology_sim2real_v2.pdf}}%
    \put(0.16074569,0.01977499){\color[rgb]{0,0,0}\makebox(0,0)[t]{\lineheight{1.25}\smash{\begin{tabular}[t]{c}\hyperref[sec_sota_system_identification]{Sytem Identification}\end{tabular}}}}%
    \put(0,0){\includegraphics[width=\unitlength,page=5]{topology_sim2real_v2.pdf}}%
    \put(0.83921415,0.16045818){\color[rgb]{0,0,0}\makebox(0,0)[t]{\lineheight{1.25}\smash{\begin{tabular}[t]{c}\hyperref[sec_sota_transfer_learning]{Transfer Learning}\end{tabular}}}}%
    \put(0,0){\includegraphics[width=\unitlength,page=6]{topology_sim2real_v2.pdf}}%
    \put(0.16074576,0.16045822){\color[rgb]{0,0,0}\makebox(0,0)[t]{\lineheight{1.25}\smash{\begin{tabular}[t]{c}\hyperref[sec_sota_curriculum_learning]{Curriculum Learning}\end{tabular}}}}%
    \put(0,0){\includegraphics[width=\unitlength,page=7]{topology_sim2real_v2.pdf}}%
    \put(0.83921407,0.09011604){\color[rgb]{0,0,0}\makebox(0,0)[t]{\lineheight{1.25}\smash{\begin{tabular}[t]{c}\hyperref[sec_sota_knowledge_distillation]{Knowledge Distillation}\end{tabular}}}}%
    \put(0,0){\includegraphics[width=\unitlength,page=8]{topology_sim2real_v2.pdf}}%
    \put(0.83921425,0.01977509){\color[rgb]{0,0,0}\makebox(0,0)[t]{\lineheight{1.25}\smash{\begin{tabular}[t]{c}\hyperref[sec_sota_sbi]{Simulation-Based Inference}\end{tabular}}}}%
    \put(0,0){\includegraphics[width=\unitlength,page=9]{topology_sim2real_v2.pdf}}%
    \put(0.49997974,0.01977509){\color[rgb]{0,0,0}\makebox(0,0)[t]{\lineheight{1.25}\smash{\begin{tabular}[t]{c}\hyperref[sec_sota_adaptive_control]{Adaptive Control}\end{tabular}}}}%
  \end{picture}%
\endgroup%

%% file: images/topology_dr.pdf_tex
\begingroup%
  \makeatletter%
  \providecommand\color[2][]{%
    \errmessage{(Inkscape) Color is used for the text in Inkscape, but the package 'color.sty' is not loaded}%
    \renewcommand\color[2][]{}%
  }%
  \providecommand\transparent[1]{%
    \errmessage{(Inkscape) Transparency is used (non-zero) for the text in Inkscape, but the package 'transparent.sty' is not loaded}%
    \renewcommand\transparent[1]{}%
  }%
  \providecommand\rotatebox[2]{#2}%
  \newcommand*\fsize{\dimexpr\f@size pt\relax}%
  \newcommand*\lineheight[1]{\fontsize{\fsize}{#1\fsize}\selectfont}%
  \ifx\svgwidth\undefined%
    \setlength{\unitlength}{230.82011765bp}%
    \ifx\svgscale\undefined%
      \relax%
    \else%
      \setlength{\unitlength}{\unitlength * \real{\svgscale}}%
    \fi%
  \else%
    \setlength{\unitlength}{\svgwidth}%
  \fi%
  \global\let\svgwidth\undefined%
  \global\let\svgscale\undefined%
  \makeatother%
  \begin{picture}(1,0.24406664)%
    \lineheight{1}%
    \setlength\tabcolsep{0pt}%
    \put(0,0){\includegraphics[width=\unitlength,page=1]{topology_dr.pdf}}%
    \put(0.49841978,0.1788776){\color[rgb]{0,0,0}\makebox(0,0)[t]{\lineheight{1.25}\smash{\begin{tabular}[t]{c}\hyperref[sec_sota_domain_randomization]{Domain Randomization}\end{tabular}}}}%
    \put(0,0){\includegraphics[width=\unitlength,page=2]{topology_dr.pdf}}%
    \put(0.49824368,0.03918726){\color[rgb]{0,0,0}\makebox(0,0)[t]{\lineheight{1.25}\smash{\begin{tabular}[t]{c}\hyperref[sec_sota_adaptive_dr]{adaptive}\end{tabular}}}}%
    \put(0.1521237,0.03919074){\color[rgb]{0,0,0}\makebox(0,0)[t]{\lineheight{1.25}\smash{\begin{tabular}[t]{c}\hyperref[sec_sota_static_dr]{static}\end{tabular}}}}%
    \put(0.84455816,0.03919115){\color[rgb]{0,0,0}\makebox(0,0)[t]{\lineheight{1.25}\smash{\begin{tabular}[t]{c}\hyperref[sec_sota_adversarial_dr]{adversarial}\end{tabular}}}}%
  \end{picture}%
\endgroup%

%% file: images/sketch_static_dr.pdf_tex
\begingroup%
  \makeatletter%
  \providecommand\color[2][]{%
    \errmessage{(Inkscape) Color is used for the text in Inkscape, but the package 'color.sty' is not loaded}%
    \renewcommand\color[2][]{}%
  }%
  \providecommand\transparent[1]{%
    \errmessage{(Inkscape) Transparency is used (non-zero) for the text in Inkscape, but the package 'transparent.sty' is not loaded}%
    \renewcommand\transparent[1]{}%
  }%
  \providecommand\rotatebox[2]{#2}%
  \newcommand*\fsize{\dimexpr\f@size pt\relax}%
  \newcommand*\lineheight[1]{\fontsize{\fsize}{#1\fsize}\selectfont}%
  \ifx\svgwidth\undefined%
    \setlength{\unitlength}{253.36158291bp}%
    \ifx\svgscale\undefined%
      \relax%
    \else%
      \setlength{\unitlength}{\unitlength * \real{\svgscale}}%
    \fi%
  \else%
    \setlength{\unitlength}{\svgwidth}%
  \fi%
  \global\let\svgwidth\undefined%
  \global\let\svgscale\undefined%
  \makeatother%
  \begin{picture}(1,0.35317086)%
    \lineheight{1}%
    \setlength\tabcolsep{0pt}%
    \put(0,0){\includegraphics[width=\unitlength,page=1]{sketch_static_dr.pdf}}%
    \put(0.83205576,0.18717705){\color[rgb]{0,0,0}\makebox(0,0)[t]{\lineheight{1.25}\smash{\begin{tabular}[t]{c}target domain\\(real robot)\end{tabular}}}}%
    \put(0.24467926,0.18706704){\color[rgb]{0,0,0}\makebox(0,0)[t]{\lineheight{1.25}\smash{\begin{tabular}[t]{c}source domain\\(rand. simulator)\end{tabular}}}}%
    \put(0.13313288,0.31351289){\color[rgb]{0,0,0}\makebox(0,0)[t]{\lineheight{1.25}\smash{\begin{tabular}[t]{c}train policy\end{tabular}}}}%
    \put(0.55234864,0.30107466){\color[rgb]{0,0,0}\makebox(0,0)[t]{\lineheight{1.25}\smash{\begin{tabular}[t]{c}execute policy\end{tabular}}}}%
  \end{picture}%
\endgroup%

%% file: images/sketch_adaptive_dr.pdf_tex
\begingroup%
  \makeatletter%
  \providecommand\color[2][]{%
    \errmessage{(Inkscape) Color is used for the text in Inkscape, but the package 'color.sty' is not loaded}%
    \renewcommand\color[2][]{}%
  }%
  \providecommand\transparent[1]{%
    \errmessage{(Inkscape) Transparency is used (non-zero) for the text in Inkscape, but the package 'transparent.sty' is not loaded}%
    \renewcommand\transparent[1]{}%
  }%
  \providecommand\rotatebox[2]{#2}%
  \newcommand*\fsize{\dimexpr\f@size pt\relax}%
  \newcommand*\lineheight[1]{\fontsize{\fsize}{#1\fsize}\selectfont}%
  \ifx\svgwidth\undefined%
    \setlength{\unitlength}{253.36213037bp}%
    \ifx\svgscale\undefined%
      \relax%
    \else%
      \setlength{\unitlength}{\unitlength * \real{\svgscale}}%
    \fi%
  \else%
    \setlength{\unitlength}{\svgwidth}%
  \fi%
  \global\let\svgwidth\undefined%
  \global\let\svgscale\undefined%
  \makeatother%
  \begin{picture}(1,0.38214212)%
    \lineheight{1}%
    \setlength\tabcolsep{0pt}%
    \put(0,0){\includegraphics[width=\unitlength,page=1]{sketch_adaptive_dr.pdf}}%
    \put(0.2446816,0.21684209){\color[rgb]{0,0,0}\makebox(0,0)[t]{\lineheight{1.25}\smash{\begin{tabular}[t]{c}source domain\\(rand. simulator)\end{tabular}}}}%
    \put(0.83205619,0.21603886){\color[rgb]{0,0,0}\makebox(0,0)[t]{\lineheight{1.25}\smash{\begin{tabular}[t]{c}target domain\\(real robot)\end{tabular}}}}%
    \put(0,0){\includegraphics[width=\unitlength,page=2]{sketch_adaptive_dr.pdf}}%
    \put(0.13313523,0.34248423){\color[rgb]{0,0,0}\makebox(0,0)[t]{\lineheight{1.25}\smash{\begin{tabular}[t]{c}train policy\end{tabular}}}}%
    \put(0.55235009,0.33646035){\color[rgb]{0,0,0}\makebox(0,0)[t]{\lineheight{1.25}\smash{\begin{tabular}[t]{c}execute policy\end{tabular}}}}%
    \put(0,0){\includegraphics[width=\unitlength,page=3]{sketch_adaptive_dr.pdf}}%
    \put(0.55827051,0.01995526){\color[rgb]{0,0,0}\makebox(0,0)[t]{\lineheight{1.25}\smash{\begin{tabular}[t]{c}collect data\end{tabular}}}}%
    \put(0.23283601,0.01085749){\color[rgb]{0,0,0}\makebox(0,0)[t]{\lineheight{1.25}\smash{\begin{tabular}[t]{c}adapt randomization\end{tabular}}}}%
    \put(0,0){\includegraphics[width=\unitlength,page=4]{sketch_adaptive_dr.pdf}}%
  \end{picture}%
\endgroup%

%% file: images/sketch_adversarial_dr.pdf_tex
\begingroup%
  \makeatletter%
  \providecommand\color[2][]{%
    \errmessage{(Inkscape) Color is used for the text in Inkscape, but the package 'color.sty' is not loaded}%
    \renewcommand\color[2][]{}%
  }%
  \providecommand\transparent[1]{%
    \errmessage{(Inkscape) Transparency is used (non-zero) for the text in Inkscape, but the package 'transparent.sty' is not loaded}%
    \renewcommand\transparent[1]{}%
  }%
  \providecommand\rotatebox[2]{#2}%
  \newcommand*\fsize{\dimexpr\f@size pt\relax}%
  \newcommand*\lineheight[1]{\fontsize{\fsize}{#1\fsize}\selectfont}%
  \ifx\svgwidth\undefined%
    \setlength{\unitlength}{253.36146937bp}%
    \ifx\svgscale\undefined%
      \relax%
    \else%
      \setlength{\unitlength}{\unitlength * \real{\svgscale}}%
    \fi%
  \else%
    \setlength{\unitlength}{\svgwidth}%
  \fi%
  \global\let\svgwidth\undefined%
  \global\let\svgscale\undefined%
  \makeatother%
  \begin{picture}(1,0.3803783)%
    \lineheight{1}%
    \setlength\tabcolsep{0pt}%
    \put(0,0){\includegraphics[width=\unitlength,page=1]{sketch_adversarial_dr.pdf}}%
    \put(0.24467476,0.2142745){\color[rgb]{0,0,0}\makebox(0,0)[t]{\lineheight{1.25}\smash{\begin{tabular}[t]{c}source domain\\(rand. simulator)\end{tabular}}}}%
    \put(0,0){\includegraphics[width=\unitlength,page=2]{sketch_adversarial_dr.pdf}}%
    \put(0.8320611,0.2142745){\color[rgb]{0,0,0}\makebox(0,0)[t]{\lineheight{1.25}\smash{\begin{tabular}[t]{c}target domain\\(real robot)\end{tabular}}}}%
    \put(0,0){\includegraphics[width=\unitlength,page=3]{sketch_adversarial_dr.pdf}}%
    \put(0.13313293,0.34072031){\color[rgb]{0,0,0}\makebox(0,0)[t]{\lineheight{1.25}\smash{\begin{tabular}[t]{c}train policy\end{tabular}}}}%
    \put(0.35663929,0.01085752){\color[rgb]{0,0,0}\makebox(0,0)[t]{\lineheight{1.25}\smash{\begin{tabular}[t]{c}train adversary\end{tabular}}}}%
    \put(0.55234888,0.32830717){\color[rgb]{0,0,0}\makebox(0,0)[t]{\lineheight{1.25}\smash{\begin{tabular}[t]{c}execute policy\end{tabular}}}}%
  \end{picture}%
\endgroup%